\documentclass[runningheads]{llncs}

\usepackage{eccv}

\usepackage{eccvabbrv}

\usepackage{graphicx}
\usepackage{booktabs}

\usepackage[accsupp]{axessibility}  
\usepackage[colorlinks=true]{hyperref}

\usepackage{orcidlink}

\usepackage{multirow}
\usepackage{colortbl}
\usepackage{ulem}
\usepackage{wrapfig}
\usepackage{booktabs}
\usepackage{array}
\usepackage{tabulary,amsmath,amssymb,overpic,xcolor,wrapfig}
\usepackage{tabularx}
\usepackage{caption}

\definecolor{mygray}{RGB}{230,230,230}

\definecolor{mygreen}{RGB}{22,136,54}
\definecolor{myyellow}{RGB}{212, 90, 22}
\definecolor{myorange}{RGB}{245,168,0}
\definecolor{mypink}{RGB}{112, 174, 251}

\newcommand{\greenbox}[1]{\textcolor{mygreen}{{#1}}}
\newcommand{\yellowbox}[1]{\textcolor{myyellow}{{#1}}}
\newcommand{\orangebox}[1]{\textcolor{myorange}{{#1}}}
\newcommand{\pinkbox}[1]{\textcolor{mypink}{{#1}}}

\begin{document}

\title{Can 3D Vision-Language Models Truly Understand Natural Language?}

\author{Weipeng Deng\inst{1} \and
Jihan Yang\inst{1} \and
Runyu Ding\inst{1} \and
Jiahui Liu\inst{1} \and
Yijiang Li\inst{2} \and \\
Xiaojuan Qi\inst{1} \and
Edith Ngai\inst{1}}

\authorrunning{W. Deng \and J. Yang \and R. Ding \and X. Qi \and E. Ngai et al.}

\institute{The University of Hong Kong \and
Johns Hopkins University
\\
\email{\{wpdeng,jhyang,ryding,liujh,xjqi,chngai\}@eee.hku.hk}}

\maketitle
\begin{figure*}[!h]
  \centering
   \includegraphics[width=0.85\textwidth]{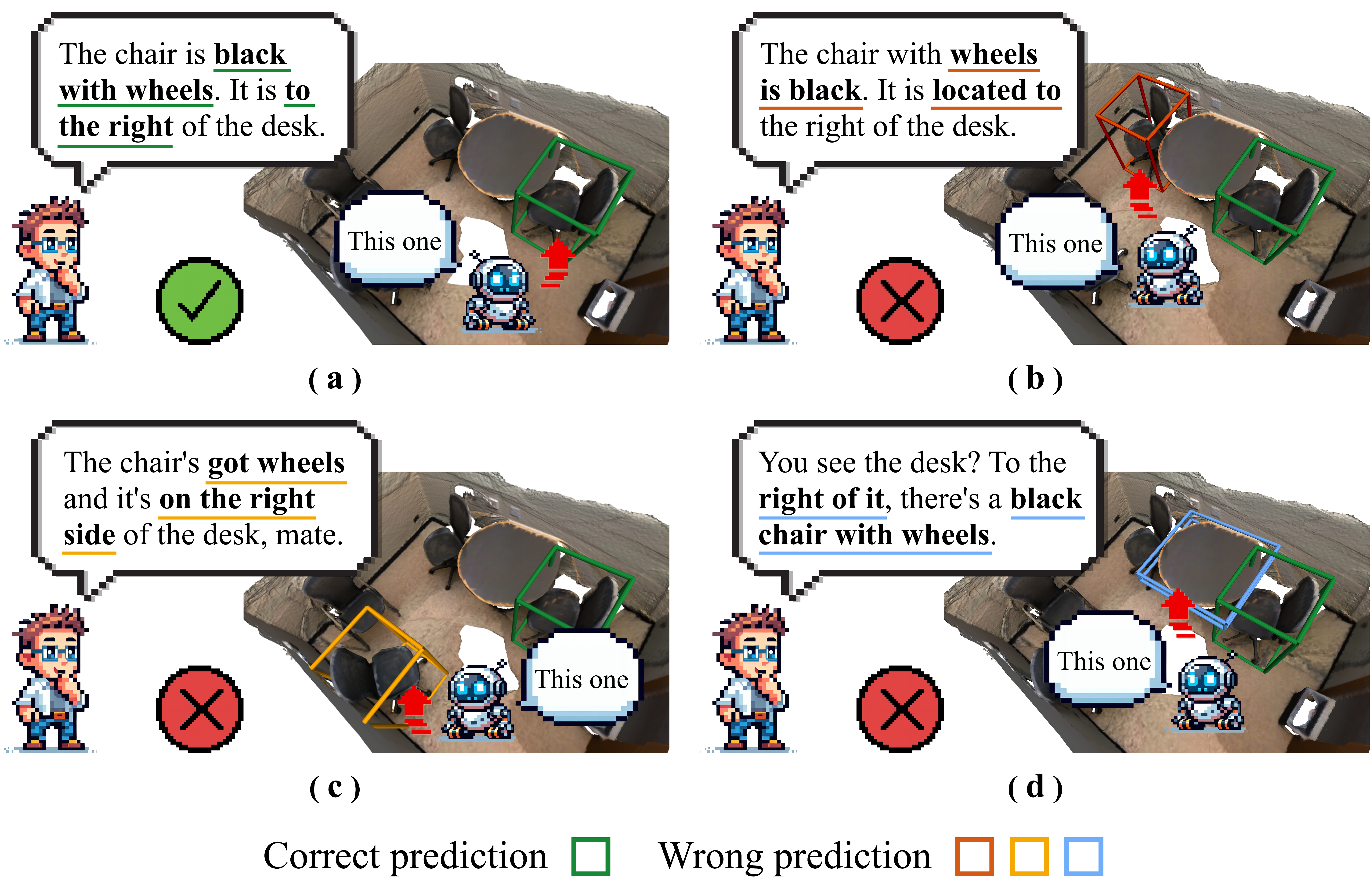}
    \vspace{-0.4cm}
   \caption{Fragility of 3D-VL models in natural language understanding. This figure illustrates the significant performance degradation of 3D Vision-Language models when faced with natural language variations common in human communication. The variations tested include: \emph{\greenbox{(a)}}. the \greenbox{original sentence} in the training set. \emph{\yellowbox{(b)}}. shifting the voice from active voice to \yellowbox{passive voice}. \emph{\orangebox{(c)}}. Saying the same thing in a \orangebox{different accent}. \emph{\pinkbox{(d)}}. Saying in a new \pinkbox{conversation tone}. These variations are common in human language, but the model fails on them.  }
   \label{fig:main_fig}
\end{figure*}
\vspace{-0.8cm}

\begin{abstract}
Rapid advancements in 3D vision-language (3D-VL) tasks have opened up new avenues for human interaction with embodied agents or robots using natural language. Despite this progress, we find a notable limitation: existing 3D-VL models exhibit sensitivity to the styles of language input, struggling to understand sentences with the same semantic meaning but written in different variants. This observation raises a critical question: \textbf{Can 3D vision-language models truly understand natural language?} To test the language understandability of 3D-VL models, we first propose a language robustness task for systematically assessing 3D-VL models across various tasks, benchmarking their performance when presented with different language style variants. Importantly, these variants are commonly encountered in applications requiring direct interaction with humans, such as embodied robotics, given the diversity and unpredictability of human language. We propose a 3D Language Robustness Dataset, designed based on the characteristics of human language, to facilitate the systematic study of robustness. Our comprehensive evaluation uncovers a significant drop in the performance of all existing models across various 3D-VL tasks. Even the state-of-the-art 3D-LLM fails to understand some variants of the same sentences. Further in-depth analysis suggests that the existing models have a fragile and biased fusion module, which stems from the low diversity of the existing dataset. Finally, we propose a training-free module driven by LLM, which improves language robustness. Datasets and code will be available at github: \url{https://github.com/VincentDENGP/3D-LR}.

\end{abstract}

\vspace{-0.5cm}
\section{Introduction}
\label{sec:intro}
\vspace{-0.2cm}
In recent years, connecting vision and language has garnered considerable interest~\cite{liu2023visual,li2023blip}. Significant progress has been made on various tasks, such as Visual Grounding and Visual Question Answering (VQA), in the context of both 2D Vision Language (2D-VL)~\cite{tiong2022plug,fukui2016multimodal,antol2015vqa,anderson2018bottom,xie2023sed,zhong2022video,gao2017video} and 3D vision-language (3D-VL) understanding~\cite{huang2022multi,yang2021sat,yang2021st3d,ding2023pla,yang2022st3d++,ding2022doda,ding2023lowis3d,yang2023regionplc}. These tasks represent the foundation abilities in real-world applications such as describing images~\cite{vinyals2015show}, embodied robotics~\cite{gupta2021embodied}, AR/VR, and autonomous agents~\cite{xi2023rise,yang2024v} that require human-machine interaction. They necessitate the model's ability to comprehend free-form natural language instructions for generating predictions. 

\begin{figure}[t]
  \centering
  \begin{subfigure}{0.24\linewidth}
    \includegraphics[width=1\linewidth]{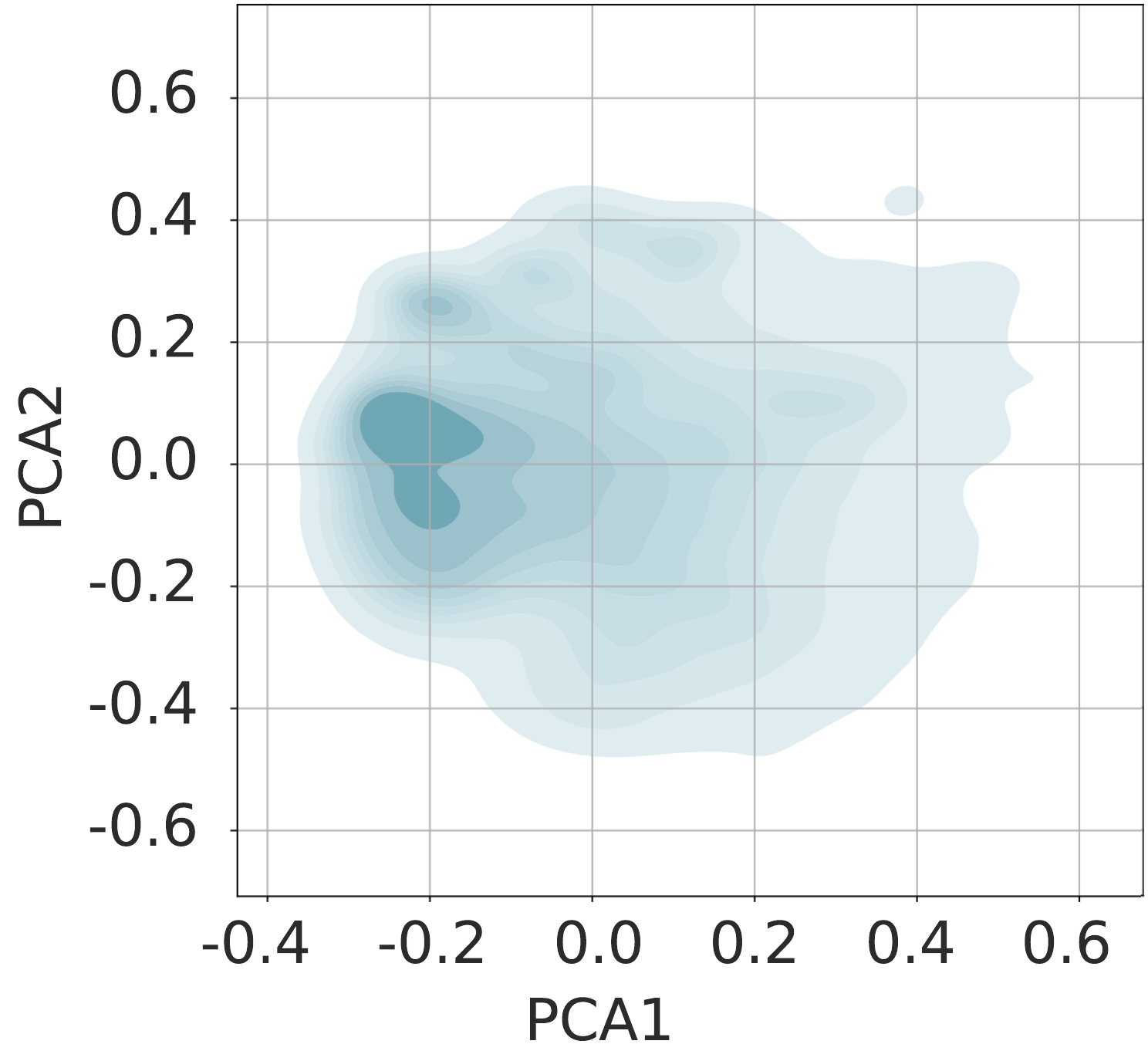}
    \caption{ScanRefer~\cite{chen2020scanrefer}}
    \label{fig:density_og}
  \end{subfigure}
  \hfill
  \begin{subfigure}{0.24\linewidth}
    \includegraphics[width=1\linewidth]{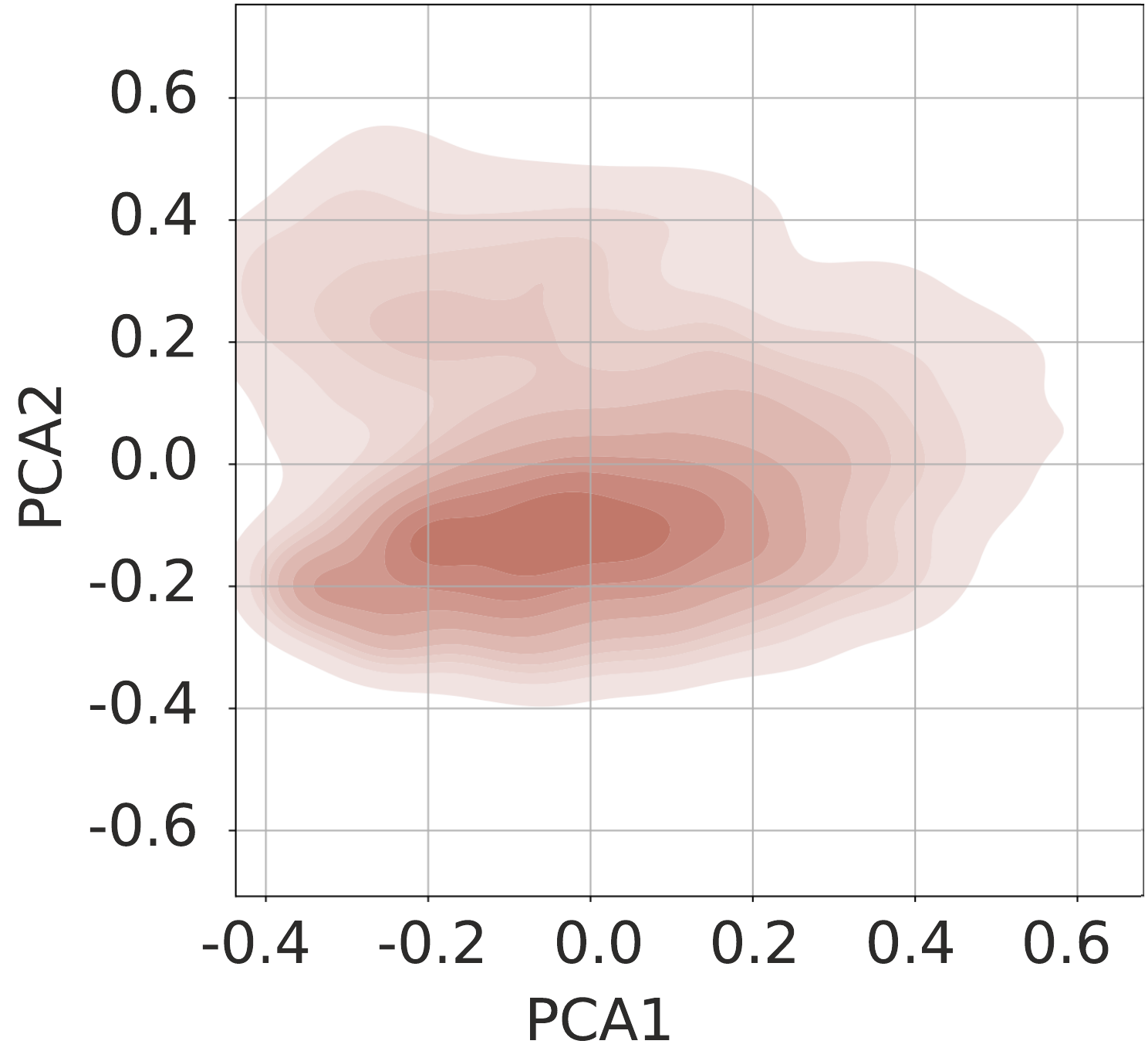}
    \caption{Our 3D-LR Dataset}
    \label{fig:density_tone}
  \end{subfigure}
  \hfill
  \begin{subfigure}{0.24\linewidth}
    \includegraphics[width=1\linewidth]{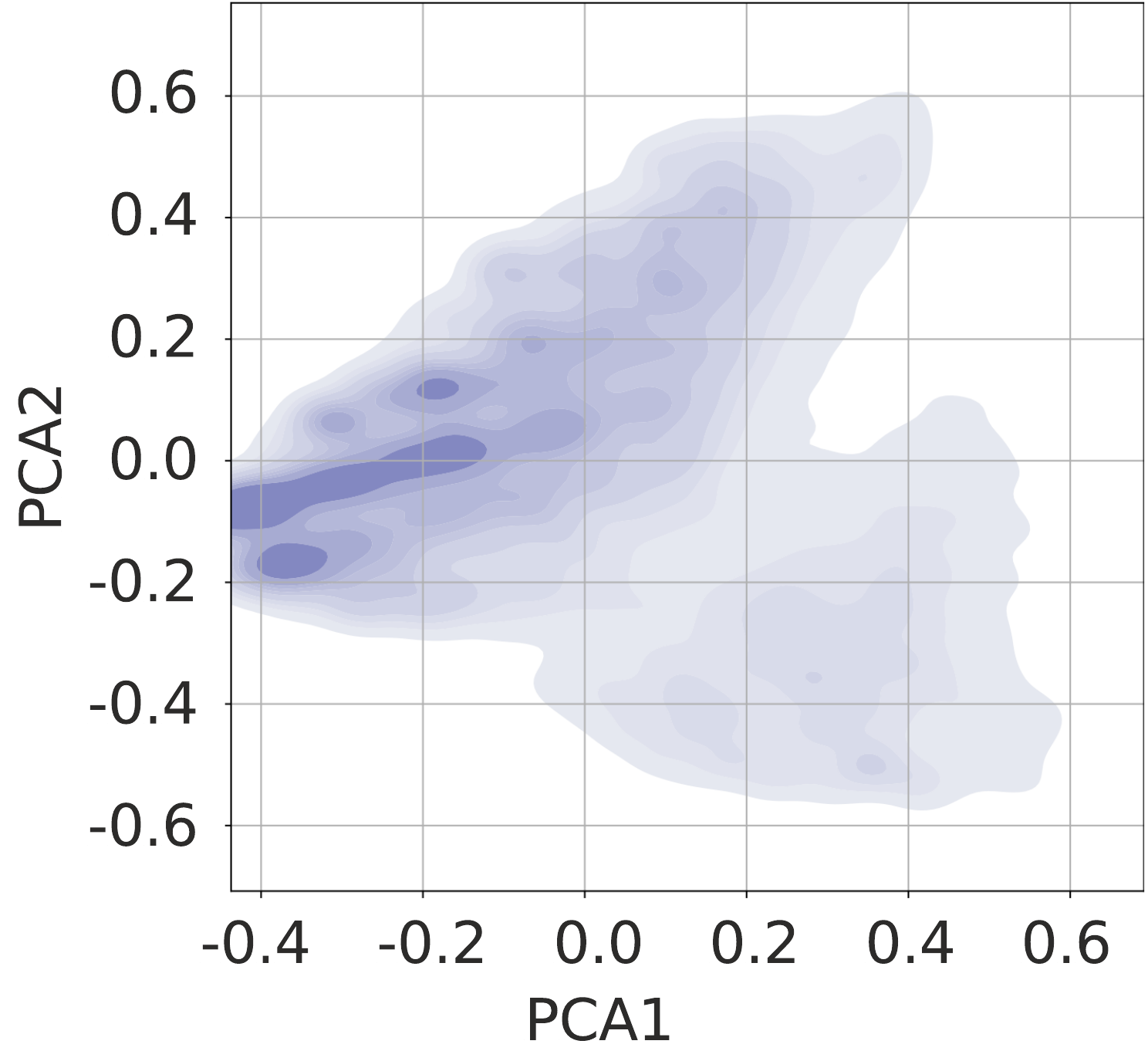}
    \caption{Coco Caption}
    \label{fig:density_coco}
  \end{subfigure}
  \begin{subfigure}{0.24\linewidth}
    \includegraphics[width=1\linewidth]{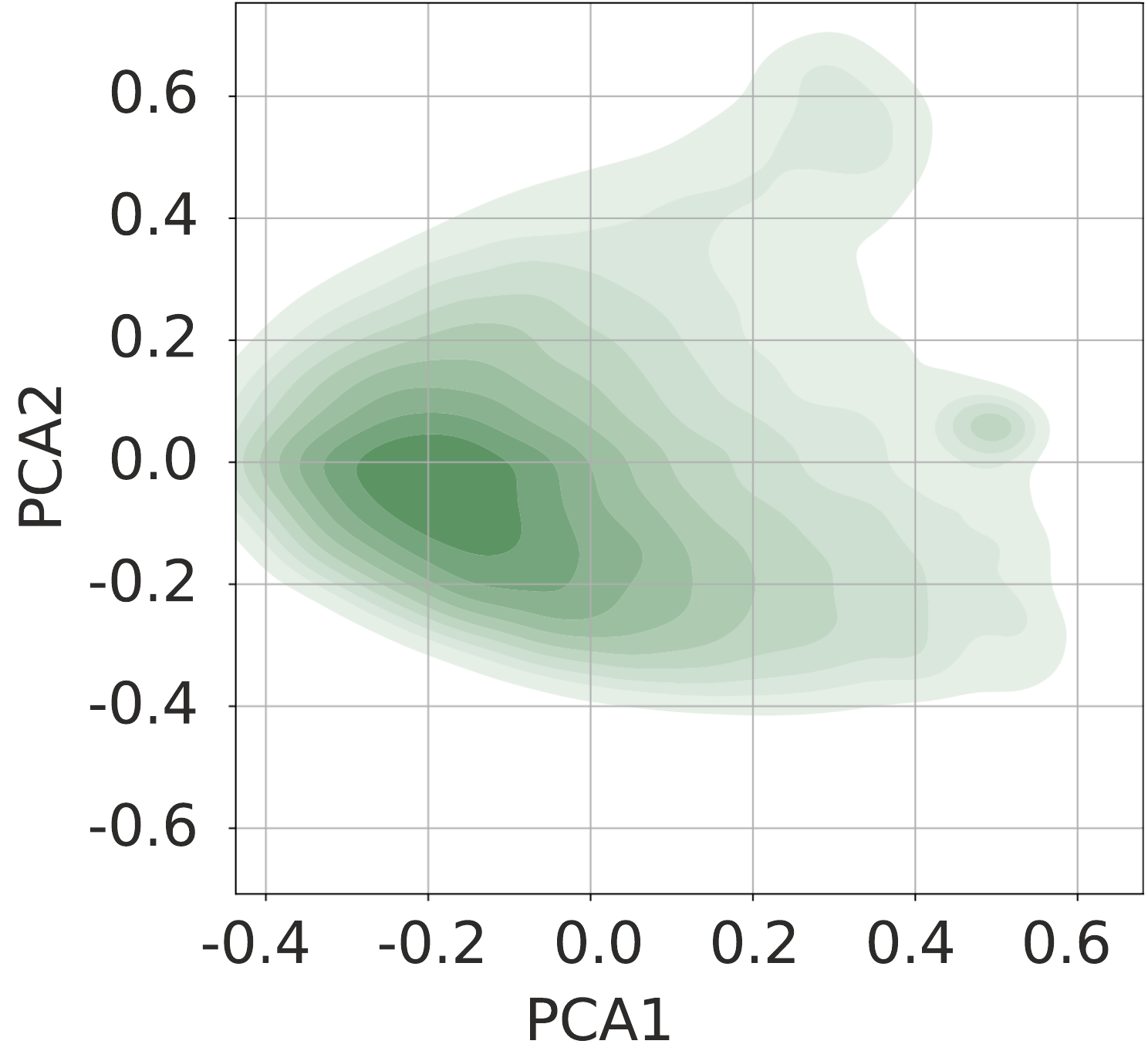}
    \caption{Open Assistant~\cite{kopf2023openassistant}}
    \label{fig:density_assiatnt}
  \end{subfigure}
  \vspace{-0.2cm}
  \caption{Density map of four datasets' vectorized syntax structure principal features. Darker areas indicate a higher concentration of similar sentence patterns. A concentrated dark region suggests that the dataset consists of simple and less diverse sentence structures. More details in Suppl.}
  \vspace{-0.8cm}
  \label{fig:densit}

\end{figure}

2D-VL models can handle various prompts~\cite{lai2023lisa}, benefiting from large-scale and diverse internet-sourced image-language datasets as shown in Fig.~\ref{fig:densit}(c). These datasets contain a wide range of natural language expressions, which enhances the robustness of 2D-VL towards different language styles. However, we didn't observe the same success in the 3D-VL domain.
Instead, we observe a notable limitation: Existing 
3D-VL models exhibit \textbf{bias towards language styles in their training datasets} and \textbf{struggle to understand minor variations} in our daily languages. As demonstrated in Fig.~\ref{fig:main_fig}, even minor variations in expression conveying the same meaning can result in model failure. Understanding different language styles is crucial for real-world applications, such as embodied robotics, where humans tend to use a variety of expressions instead of adhering to a fixed language pattern~\cite{holtzman2019curious}. 

Unfortunately, replicating the success of 2D-VL by obtaining such diverse and large-scale 3D-VL datasets is both challenging and resource-intensive, hindering the success of 3D vision-language tasks compared to their 2D counterparts. This disparity in dataset diversity and robustness motivates us to systematically study the language robustness of 3D-VL models and explore methods to improve them without relying on extensive datasets. Currently, it lacks a suitable task or dataset designed to facilitate this line of study. Recent studies have assessed 2D-VL models' robustness using negative samples like semantically altered instructions~\cite{yuksekgonul2022and,hendricks2021probing,wang2023can,thrush2022winoground,zhao2022vl} and evaluated LLMs' resilience to typo errors~\cite{liang2022holistic} (semantic preserving), which diverges from our research focus. Semantic alterations compromise the objectives of grounding or QA by distorting meaning, while simple typos fail to capture the systematic diversity of human language. Besides, we focus on studying model robustness toward \textbf{natural variations} of sentences without altering their meanings, which is more practical for real-world applications in embodied agents and robotics.

Thus, we introduce the 3D Language Robustness (3D-LR) Benchmark, designed for a comprehensive evaluation of the language robustness in 3D-VL models. 
Specifically, our benchmark evaluates various 3D-VL models on different tasks~\cite{achlioptas2020referit3d,yang2021sat,chen2020scanrefer,huang2022multi,azuma2022scanqa}, utilizing a specially curated 3D Language Robustness dataset. This dataset challenges models with a variety of language style variants.
To accurately model human natural language, we first identify the five most common key language styles variants of natural language in human communications: syntax, voice, modifier, accent, and tone, drawing from established linguistic theories~\cite{barber_beal_shaw_2009,bhagat2013paraphrase}. Each variant corresponds to a specific aspect of language commonly used in human communication. For example, syntax involves altering sentence structures, while voice entails transitioning between active and passive forms. (More details will be shown in Sec.~\ref{sec:sec3}) 
Subsequently, we developed a paraphrasing pipeline leveraging a Large Language Model~\cite{raffel2020exploring,brown2020language,wei2021finetuned,sanh2021multitask,ouyang2022training} (LLM) to generate the 3D-LR dataset. This process involves rewriting sentences from existing 3D-VL~\cite{achlioptas2020referit3d,chen2020scanrefer,azuma2022scanqa} datasets. We prompt the LLM with strict rules and linguistic theory, instructing it to rephrase sentences into designated language styles while preserving their original meaning. Furthermore, we employ both statistical analyses and neural-based semantic quality assessments to verify the variants preserve the same meaning. The evaluation results reveal that even state-of-the-art methods struggle with minor sentence style changes, experiencing performance decreases of up to 32\%. Notably, powerful LLM-based methods, such as 3D-LLM~\cite{hong20233d}, whether fine-tuned with task-specific supervision or not, also exhibit performance degradation. 

Apart from our 3D-LR benchmark, we further propose a simple LLM-based pre-alignment module that enhances robustness in 3D-VL models without additional training. Our method successfully narrows the performance gap by up to 80\%. Remarkably, it performs as well as models augmented with double training data size (from 40k to 80k). Through our comprehensive analysis, we have identified that the fusion module in existing models acts as a major point of failure, as it is biased toward the training dataset. Our proposed method can effectively address this issue. 

In summary, our primary contributions are:
\begin{itemize}
	\item [1)]
        {We present the study exploring language robustness toward natural variations of sentences without altering their meanings, which is more practical for real-world applications. We aim to answer an important question: \textbf{Can 3D vision-language models truly understand natural language?}}
        \item [2)]
        {We conduct a systematically designed 3D language robustness dataset based on linguist theories, which properly models real-world natural language to facilitate system benchmarking. Our benchmarks on various 3D-VL models revealed their vulnerability to language patterns. Further in-depth analysis showed that this issue stems from the fusion module, primarily caused by the limited diversity in the training datasets.}
	\item [3)]
        {We propose a simple yet effective training-free LLM-based pre-alignment module that can recover a large proportion of performance without training.}
\end{itemize}

\vspace{-0.5cm}
\section{Related Works}
\begin{figure}[htbp]
  \centering
  \vspace{-1.1cm}
  \begin{subfigure}{0.45\linewidth}
  \centering
    \includegraphics[height=3cm,keepaspectratio]{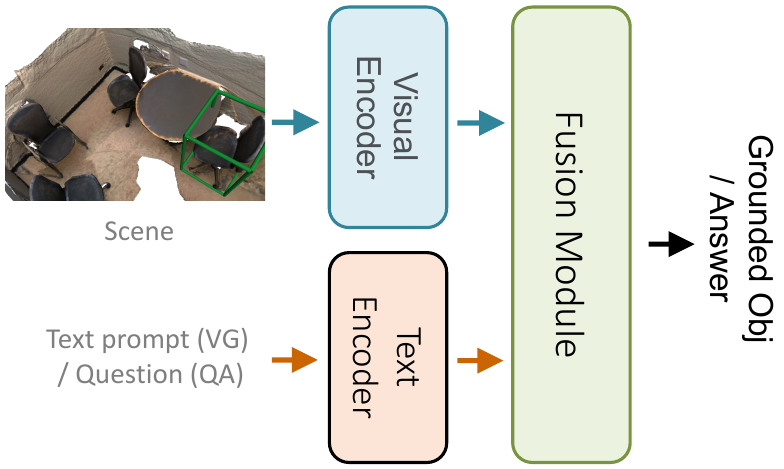}
    \caption{Dual stream architecture of 3D-VL. Each branch encodes a modality; the fusion model aligns features for prediction.}
    \label{fig:arch1}
  \end{subfigure}
  \hfill
  \begin{subfigure}{0.45\linewidth}
  \centering
    \includegraphics[height=3cm,keepaspectratio]{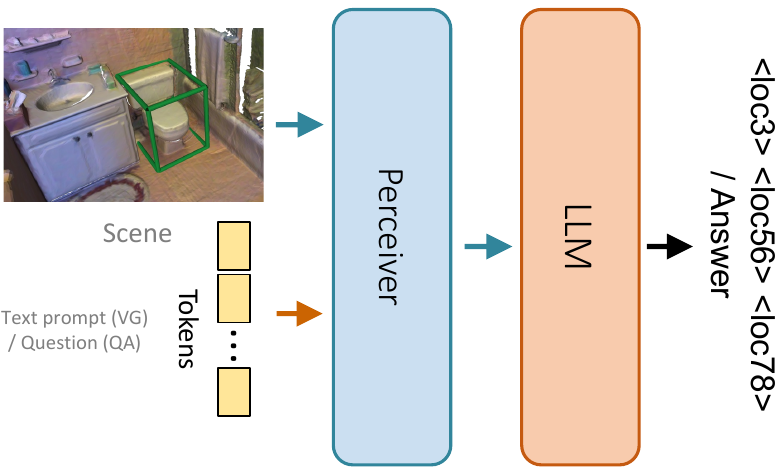}
    \caption{LLM-based architecture of 3D-VL. Using 3D-LLM as an example~\cite{hong20233d}: The perceiver encodes and aligns two modalities, while the LLM processes the output. }
    \label{fig:arch2}
  \end{subfigure}
  \vspace{-0.2cm}
  \caption{Review of popular model architectures in 3D-VL. (a) illustrates traditional dual-stream models, while (b) presents the recently proposed LLM-based architecture.}
  \vspace{-0.8cm}
  \label{fig:archs}
\end{figure}

\noindent\textbf{3D Vision-Language (3D-VL).} 
3D-VL understanding tasks, such as 3D Visual Question Answering (3D-VQA) and 3D Visual Grounding (3D-VG), are pivotal for embodied agents and robotics. These tasks require the simultaneous perception of the 3D world and an understanding of natural language. In 3D-VQA, models select the correct answer from a set of candidates based on the input 3D scene and a natural language question~\cite{azuma2022scanqa} while 3D-VG involves selecting the correct object~\cite{achlioptas2020referit3d,chen2020scanrefer}. 

The key to both tasks is aligning 3D and text, leading them to have similar model architectures. The most common architectural design is a dual streams model~\cite{azuma2022scanqa,zhang2023multi3drefer,ye20223d,ma2022sqa3d,achlioptas2020referit3d,abdelreheem20223dreftransformer,chen2020scanrefer}. This comprises a visual encoder~\cite{qi2017pointnet}, and a text encoder (LSTM~\cite{hochreiter1997long} or BERT~\cite{Devlin2019BERTPO}) that converts natural language. Then a fusion module aligns features before passing them to the prediction head, as shown in Fig.~\ref{fig:archs}(a). Additionally, leveraging multi-view 2D images has shown promise in enhancing visual feature representation (MVT~\cite{huang2022multi}, SAT~\cite{yang2021sat}). Inspired by the 2D vision large language model pre-training, there are some attempts at bringing LLM into 3D-VL~\cite{hong20233d}. As shown in Fig.~\ref{fig:archs}(b), this line of work adapts 3D features into language space, enabling the LLM to use visual features as conditions to make predictions.

Existing 3D-VL datasets are constructed either via computational methods~\cite{achlioptas2020referit3d} or created by human~\cite{achlioptas2020referit3d,chen2020scanrefer}.
There is ongoing debate about whether these datasets accurately model 3D-VL tasks. SQA3D~\cite{ma2022sqa3d} highlights the necessity for an embodied agent to know its current situation before performing tasks. They introduced the SQA3D dataset includes situation descriptions. Multi3Drefer~\cite{zhang2023multi3drefer} challenges the assumption in existing 3D-VG tasks that only one object is referenced in a sentence. Despite these advancements, we observe a noticeable ``{domain gap}'' between the language used in these datasets and everyday human language. Specifically, even datasets labeled by humans lack diversity in language style, showing a quite fixed pattern (Fig.~\ref{fig:densit}(a)), which is different from the varied expressions found in natural human communication (Fig.~\ref{fig:densit}(d)). Therefore, we propose a language robustness task with an evaluation dataset to assess the models' ability to interpret sentences in various styles.

\vspace{0.1in}\noindent\textbf{Model Robustness Studies.}
Several studies have explored visual robustness in 2D~\cite{hendrycks2019benchmarking} and 3D~\cite{ren2022pointcloud}. For language robustness, there are two categories: semantic alteration and semantic preservation. 1) \textbf{Semantic alteration} researches modifies specific textual elements in prompts, such as rearranging phrases or swapping attributes~\cite{yuksekgonul2022and,hendricks2021probing,wang2023can,thrush2022winoground,zhao2022vl}, to learn models' behavior under varying textual conditions. This approach is unsuitable for the 3D-VL setting. Core tasks in 3D-VL, such as 3D-VG and 3D-VQA, demand high semantic accuracy. Such alteration changes the textual meaning which contradicts the task's goal. Besides, we aim to explore how models handle the diversity of human expression, rather than creating unnatural test sets to challenge the model. On the other hand, 2) \textbf{Semantic preservation} studies investigate how LLMs manage minor typos~\cite{liang2022holistic}. This study is only conducted within the Natural Language Processing (NLP) unimodality domain. Besides, it limits its scope to simple spelling modifications and overlooks the complex and systematic variations in human language that our research aims to model. Our work falls within the semantic preservation category, distinct from studies focused on typos. Our research delves into the complex, systematic, and practical natural variations in language expression.

\section{3D Language Robustness (3D-LR) Benchmark}
\label{sec:sec3}
3D-VL systems aim to use natural language as an interface between humans and intelligence agents/robots. These systems should be capable of handling arbitrary human instruction style since humans' natural language is arbitrary and creative~\cite{barber_beal_shaw_2009,bhagat2013paraphrase}.
However, we observed that existing 3D-VL models are vulnerable and fail when the input text contains minor stylistic variations. Fig.~\ref{fig:main_fig} shows some examples using 3D Visual Grounding (3D-VG). When the input sentence \uline{``The chair is black with wheels. It is to the right of the desk.''} is rephrased to \uline{``The chair with wheels is black. It is located to the right of the desk.''} A SOTA grounding model, SAT~\cite{yang2021sat}, fails on prediction. These two sentences share the same semantic meaning, which is easy for humans to handle. This observation raises a natural question: \textbf{Can 3D vision-language models truly understand natural language?} It motivates us to study the language robustness problem of existing 3D-VL models, aiming to gain a systematic assessment of different models and tasks. 

To systematically evaluate the fragility of 3D-VL methods to various language styles, we propose the 3D Language Robustness benchmark. The subsequent sections detail our benchmark design, starting with our proposed 3D Language Robustness Task in Sec.~\ref{sec3:task}. Following this, we introduce our 3D Language Robustness dataset (3D-LR) in Sec.~\ref{sec3:ds}, including its design principles, construction pipeline, and key statistics.

\subsection{3D Language Robustness Task}
\label{sec3:task}
Motivated by the above, we present the 3D Language Robustness task. This task is designed to evaluate the generalization capabilities of a pre-trained 3D-VL model across diverse language variants of a given dataset. Specifically, this evaluates the model's ability to understand and process sentences that have the same meaning as the original but are expressed differently.

Formally, a standard 3D-VL task such as 3D-VQA~\cite{azuma2022scanqa}, 3D-VG~\cite{chen2020scanrefer} can be viewed as the model takes two modality inputs: a 3D scene represent in a $k$ points point cloud $\mathcal{P}=\{(p_{i},f_{i}); i = 1, 2, \ldots, k\}$, where $p_{i} \in \mathcal{R}^3$ denotes the coordinates and $f_{i}$ represents extra features and a natural language sentence $\mathcal{S} = [s_{0}, s_{1}, ..., s_{n}]$ representing a free-from sentence with $n$ words. The model treats these tasks as a classification problem over a predefined set of candidate answers or objects.

To simulate daily natural language variants, we first derived the five most representative language characteristics that humans use in communication from linguistic theory~\cite{barber_beal_shaw_2009,bhagat2013paraphrase}. Based on these characteristics, we design five rephrasing operators $o \in \{N=5 \text{ styles}\} $ for translating the original 3D-VQA and 3D-VG datasets,$\mathcal{D}$, into various styles while maintaining the meaning. Resulting in different sets of dataset splits, denoted as $\mathcal{D}_{o}$, containing sentence variants respectively. These different language variants contain the same information as the original sentences. They include the very same keywords. In other words, all language variants share the same language clues to finish the task.
Given such a series of textual variant splits represent the same meaning. A model, denoted as $\mathcal{M}$ trained on the original data $\mathcal{D}$ is evaluated on the newly built data splits $\mathcal{D}_{o}$ as described above. We have five language variant splits in our setting. Aiming to evaluate the performance degradation comprehensively.

\subsection{3D Language Robustness (3D-LR) dataset}
\label{sec3:ds}

\vspace{0.05in}\noindent\textbf{Background of Language Characteristics.}
Human language conveys meaning through its flexible syntax, grammar structures, and features like voice and tones. We use \textbf{five} main characteristics to model human language inspired by linguist theories~\cite{barber_beal_shaw_2009,bhagat2013paraphrase}. Firstly, \textbf{Syntax} refers to varying word or phrase orders to create different sentence structures; for example, inverse sentences are commonly used in daily conversation. Secondly, \textbf{Voice} involves paraphrasing a sentence from active to passive voice, or vice versa, a fundamental aspect of human language. Thirdly, \textbf{Modifier}, such as adjectives and adverbs, are varied by humans to enhance the details in a sentence, adding richness and depth to communication. Fourthly, \textbf{Accent} reflects the distinct linguistic habits of English speakers from different regions, characterized by unique vocabulary and sentence structures. These regional variations, however, do not change the fundamental meaning of the communication. Finally, \textbf{Tone} encompasses the attitudes and emotions conveyed in a sentence, which vary across different contexts. An example is the use of questions in daily conversation, which demonstrates how tone can add layers of meaning beyond the literal interpretation of words. Some examples are shown in Fig.~\ref{fig:prompt_example}. A detailed explanation is shown in the supplementary file.

\begin{wrapfigure}[19]{r}{0.48\textwidth}
    \vspace{-20pt}
    \centering
    \includegraphics[width=1\linewidth]{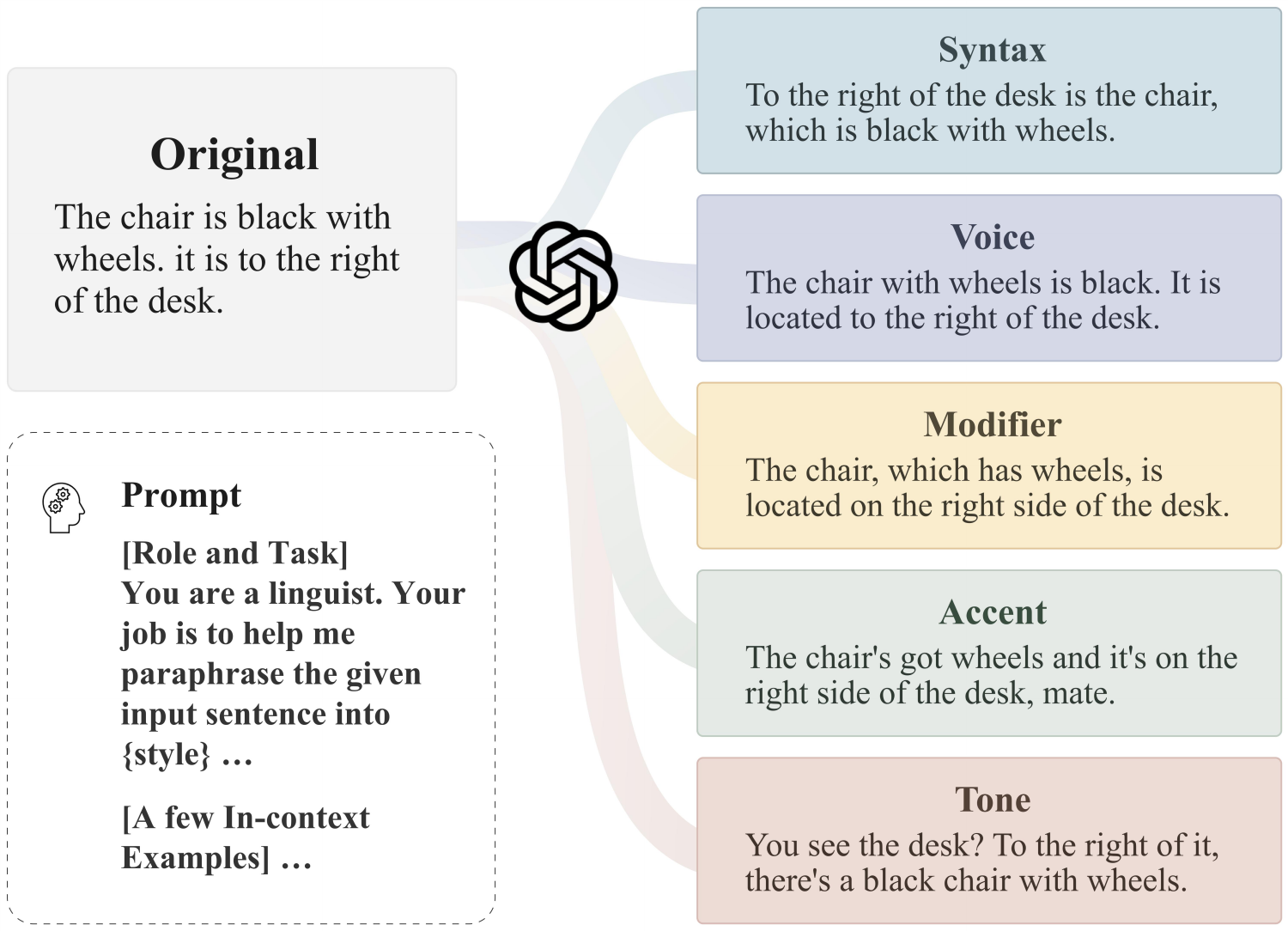}
    \caption{The overall rephrase process to build our proposed evaluation suite and the abstract prompt design. We prompt \textit{gpt} to paraphrase the original sentence into five different styles derived from human natural language characteristics. One example from ScanRefer~\cite{chen2020scanrefer} is shown in different styles. }
   \label{fig:prompt_example}
\end{wrapfigure}

\vspace{0.15in}\noindent\textbf{LLM Rephrasing for Dataset Construction.}

We aim to create a dataset including the language variants defined earlier to systematically assess the language robustness of existing models across different linguist styles. Our approach involves rephrasing sentences from existing datasets: ScanQA~\cite{azuma2022scanqa} for 3D-VQA, NR3D~\cite{achlioptas2020referit3d} and ScanRefer~\cite{chen2020scanrefer} for 3D-VG. These rephrased sentences adhere to our variant definitions while retaining their original meaning. Formally, let $\mathcal{D} \in \{\text{ScanQA}, \text{NR3D}, \text{ScanRefer}\}$ and $o \in \{\text{Syntax}, \text{Voice},  \text{Modifier}, \text{Accent}, \\ \text{Tone}\} $.

We utilize the large language model (LLM)-\textit{gpt-3.5-turbo}, for this sentence rephrasing task. Our methodology involves designing specific prompts, incorporating structured prompting and Chain of Thought (CoT~\cite{wei2022chain}) techniques to guide paraphrasing. An abstract version of this prompt, as shown in Fig.~\ref{fig:prompt_example}, begins with a detailed and precise definition of variant derived from linguistic theories, represented as ``{style}'' in Fig.~\ref{fig:prompt_example}. This step aims to familiarize the LLM with the desired style. Following this, we provide three pairs of human-written, in-context examples as seeds to help the LLM understand the source style and achieve better rephrasing quality. The complete version of our prompt is available in the supplementary file. 

To manage costs, we only utilize 25\% of each dataset's data. For 3D-VG datasets ScanRefer and NR3D, we sub-sample from the official test split. And for ScanQA, we sample from the validation split. We employ uniform sampling for the NR3D subset from Referit3D~\cite{achlioptas2020referit3d}, which categorizes difficulty levels based on the number of object nouns in the natural language sentences and offers view-dependent and view-independent splits for more refined assessment. This ensures that our 25\% subset maintains the integrity of the original data distribution.

\begin{table*}[t]
\centering
\caption{
Dataset quality assessment. We calculate the semantic similarity between the original ScanRefer and our paraphrased one, named \textit{ScanRefer-R}, in (a). Additionally, we present the basic statistics of our \textit{ScanRefer-R} in (b). These ensure our paraphrases maintain high semantic integrity.
}
\vspace{-0.2cm}
\label{tab:merge_stastics}

\begin{minipage}[b]{0.47\linewidth}
\centering
\caption*{\textbf{(a)} Semantic similarity of our \textit{ScanRefer-R} and the original data regarding cosine similarity of BERT embedding, Glove embedding, and Edit Distance (ED). This suggests that our data effectively retains the original's meaning.}

\vspace{-0.3cm}
\centering

\scalebox{0.75}{
\begin{tabular}{cccc}
      \toprule
      & BERT Sim. ($\uparrow$) & Glove Sim. ($\uparrow$) & ED ($\downarrow$) \\
      \midrule
      Syntax & 0.92 & 0.98 & 0.42 \\
      Voice & 0.92 & 0.98 & 0.60 \\
      Modifier & 0.91 & 0.96 & 0.51 \\
      Accent & 0.83 & 0.89 & 0.63 \\
      Tone & 0.86 & 0.93 & 0.51 \\
      \bottomrule
    \end{tabular}%
    \label{tab:semantic_sim}
}
\end{minipage}
\hfill
\begin{minipage}[b]{0.48\linewidth}
\centering
\caption*{\textbf{(b)} 
Basic statistics of different data splits in \textit{ScanRefer-R}. We calculate the number of unique words, the total number of words, and the average sentence length in each split. Statistics of other splits are shown in Suppl..
}
\vspace{-0.3cm}
\centering

\scalebox{0.65}{
\begin{tabular}{cccc}
\hline
\toprule
& Unique words & Total words & Avg. description length \\

\midrule
Original & 1254 & 42928 & 18.06 \\
\hline 
Syntax & 1212 & 38322 & 16.12 \\
Voice & 1216 & 39603 & 16.66 \\
Modifier & 1273 & 42011 & 17.67 \\
Accent & 1343 & 50533 & 21.26 \\
Tone & 1175 & 48726 & 20.50 \\ 
\bottomrule
\end{tabular}%
\label{tab:ds_stastics}
}
\end{minipage}
\vspace{-0.8cm}
\end{table*}


\vspace{0.05in}\noindent\textbf{Basic Statistics and Quality Assessments.}

We build the \textbf{3D} \textbf{L}anguage \textbf{R}obustness dataset, namely \textbf{3D-LR}, for our systematic benchmark. 3D-LR covers 2 tasks: \textit{ScanRefer-R}, \textit{NR3D-R} for 3D-VG and \textit{ScanQA-R} for 3D-VQA. Within each task split, there are six different subsets, covering five language variants and one original version without rephrasing for comparison.

\vspace{0.1in}\noindent\textbullet\ \textit{ Basic Statistics.}
Our \textit{ScanRefer-R} has 2377 sentences, \textit{NR3D-R} includes 1870 utterances, and \textit{ScanQA-R} contains 1168 questions. \cref{tab:ds_stastics}(b) presents basic statistics of \textit{ScanRefer-R}, comparing original sentences to their rephrased versions across different characteristics.
Our primary focus is on simulating diverse grammatical expression structures rather than lexical richness. To this end, we implemented strict rules ensuring that paraphrasing does not alter the meaning or core object nouns. This approach resulted in similar unique words across the different dataset splits.
In the accent and tone split, aiming to emulate a conversational style, we added some verbal phrases to the original sentences. This modification increased the average length of descriptions. 

To further analyze the diversity characteristic of our dataset and its comparators, we visualized the syntax diversity of our dataset using vectorized syntax trees that represent sentence structures. (Fig.~\ref{fig:densit}(b)). First, we extract the sentence structures into syntax trees. Then, we vectorize these trees and perform Principal Component Analysis~\cite{pearson1901liii} (PCA) to project them onto a 2-dimensional space. Finally, we create a density map based on the features. Before creating the density map, we combined all five variant splits.
For comparison, we analyzed a fully human-annotated NLP dataset, the Open Assistant~\cite{kopf2023openassistant}, which models human conversation (Fig.~\ref{fig:densit}(d)). The density maps of both datasets exhibit similar spreading patterns and area sizes. This similarity suggests that our dataset accurately reflects the diversity of human language.

\vspace{0.1in}\noindent\textbullet\ \textit{Dataset Quality Assessment.} To ensure the quality of rephrasing the dataset to facilitate a fair assessment, we further evaluate the rephrased dataset's quality centers on preserving original sentence meanings. We employ traditional metrics and language model methods to check this. Specifically, we measure the average edit distance~\cite{ristad1998learning} between each original and rephrased sentence, with results presented in \cref{tab:semantic_sim}(a). A smaller edit distance value signifies a closer sentence structure, indicating minimal alteration. Additionally, we assess semantic integrity using cosine similarity between sentence vectors derived from a neural language model BERT~\cite{Devlin2019BERTPO} and Glove~\cite{pennington2014glove} of the original and rewritten sentences. Higher scores suggest a greater preservation of semantic content. More explanations of the quality assessment method are detailed in the Supplementary file. Our findings, underscored by manual review, reveal that the rephrased dataset largely maintains its semantic essence. Notably, the observed reduction in cosine similarity scores below 0.9, particularly in accent-based sentence rephrasing, can be attributed to adding some spoken expressions and phrases (such as ``Hey bro''). After humans verify, the underlying meaning remains.

\vspace{-0.3cm}
\section{Experiments}
\label{sec:manuscript}

\begin{table*}[t]
\centering
\caption{Experimental results on 3D-VG tasks with/without our pre-alignment module. The best results are in \textbf{bold}. ORACLE refers to the original dataset performance.}
\label{tab:merge_results}
\vspace{-0.2cm}
\begin{minipage}[b]{\linewidth}
\centering
\caption*{\textbf{(a)} Results of 3D-VG task with predicted proposal. Evaluating ScanRefer and MVT with/without our module on ScanRefer~\cite{chen2020scanrefer} dataset. Performance measured in accuracy@kIoU. More explanation about the metric is in the supplementary.}
\centering
\resizebox{\textwidth}{!}{%
\begin{tabular}{cccccccccccc}
\hline
\multirow{2}{*}{Backbone} & \multicolumn{1}{c|}{\multirow{2}{*}{Method}} & \multicolumn{2}{c}{Syntax} & \multicolumn{2}{c}{Voice} & \multicolumn{2}{c}{Modifier} & \multicolumn{2}{c}{Accent} & \multicolumn{2}{c}{Tone} \\
 & \multicolumn{1}{c|}{} & Acc@0.25 & Acc@0.5 & Acc@0.25 & Acc@0.5 & Acc@0.25 & Acc@0.5 & Acc@0.25 & Acc@0.5 & Acc@0.25 & Acc@0.5 \\ \hline
\rowcolor{mygray}
\multicolumn{12}{c}{ORACLE: (Acc@0.25 = 42.36       Acc@0.5 = 27.68)} \\ \hline
\multirow{2}{*}{ScanRefer~\cite{chen2020scanrefer}} & \multicolumn{1}{c|}{baseline} & 11.32 & 7.66 & 19.73 & 13.50 & 17.04 & 11.49 & 12.79 & 8.79 & 9.55 & 6.86 \\
 & \multicolumn{1}{c|}{w. ours} & \textbf{24.95} & \textbf{16.45} & \textbf{22.17} & \textbf{14.35} & \textbf{21.33} & \textbf{14.77} & \textbf{24.48} & \textbf{15.90} & \textbf{26.42} & \textbf{17.29} \\ \hline
\rowcolor{mygray}
\multicolumn{12}{c}{ORACLE: (Acc@0.25 = 41.27        Acc@0.5 = 33.74)} \\ \hline
\multirow{2}{*}{MVT~\cite{huang2022multi}} & \multicolumn{1}{c|}{baseline} & 28.99 & 23.85 & 31.09 & 25.70 & 33.66 & 27.89 & 38.12 & 31.76 & 29.70 & 24.57 \\
 & \multicolumn{1}{c|}{w. ours} & \textbf{38.50} & \textbf{31.89} & \textbf{34.58} & \textbf{28.36} & \textbf{37.53} & \textbf{30.96} & \textbf{40.05} & \textbf{33.40} & \textbf{38.66} & \textbf{31.85} \\ \hline
\end{tabular}%
\label{tab:scanrefer_box}
}
\end{minipage}
\begin{minipage}[b]{0.54\linewidth}
\centering
\vspace{0.3cm}
\caption*{\textbf{(b)} Results of 3D-VG task with GT proposal on the NR3D~\cite{achlioptas2020referit3d} dataset. Follow the setting of NR3D, a 3D-VG task using GT box as object proposal. We measure listening accuracy~\cite{achlioptas2020referit3d} of Referit3D and SAT with/without our module.}
\centering
\resizebox{\linewidth}{!}{
\begin{tabular}{ccccccc}
\hline
Backbone & \multicolumn{1}{c|}{Method} & Syntax & Voice & Modifier & Accent & Tone \\ \hline
\rowcolor{mygray}
\multicolumn{7}{c}{ORACLE: (Acc = 35.1)} \\ \hline
\multirow{2}{*}{Referit3D~\cite{achlioptas2020referit3d}} & \multicolumn{1}{c|}{baseline} & 25.7 & 22.1 & 24.0 & 27.9 & 21.7 \\
 & \multicolumn{1}{c|}{w. ours} & \textbf{28.1} & \textbf{28.5} & \textbf{28.1} & \textbf{30.4} & \textbf{33.7} \\ \hline
 \rowcolor{mygray}
\multicolumn{7}{c}{ORACLE: (Acc = 48.8)} \\ \hline
\multirow{2}{*}{SAT~\cite{yang2021sat}} & \multicolumn{1}{c|}{baseline} & 43.9 & 34.2 & 38.9 & 41.2 & 34.9 \\
 & \multicolumn{1}{c|}{w. ours} & \textbf{45.1} & \textbf{44.1} & \textbf{41.6} & \textbf{44.6} & \textbf{46.9} \\ \hline
\end{tabular}%
\label{tab:NR3D_gt}
}
\end{minipage}
\hfill
\begin{minipage}[b]{0.43\linewidth}
\centering
\caption*{\textbf{(c)} Results of 3D-VG task with GT proposal on ScanRefer~\cite{chen2020scanrefer}. Measuring listening accuracy of Referit3D and SAT that trained on NR3D and ScanRefer, aligned with NR3D.}
\centering
\resizebox{\linewidth}{!}{
\begin{tabular}{ccccccc}
\hline
Backbone & \multicolumn{1}{c|}{Method} & Syntax & Voice & Modifier & Accent & Tone \\ \hline
\rowcolor{mygray}
\multicolumn{7}{c}{ORACLE: (Acc = 34.8)} \\ \hline
\multirow{2}{*}{Referit3D~\cite{achlioptas2020referit3d}} & \multicolumn{1}{c|}{baseline} & 16.8 & 23.9 & 24.0 & 24.0 & 17.9 \\
 & \multicolumn{1}{c|}{w. ours} & \textbf{31.9} & 23.2 & \textbf{24.5} & \textbf{27.2} & \textbf{34.4} \\ \hline
 \rowcolor{mygray}
\multicolumn{7}{c}{ORACLE: (Acc = 40.7)} \\ \hline
\multirow{2}{*}{SAT-NR3D~\cite{yang2021sat}} & \multicolumn{1}{c|}{baseline} & 37.1 & 33.2 & 36.6 & 34.7 & 33.1 \\
 & \multicolumn{1}{c|}{w. ours} & \textbf{39.7} & \textbf{35.7} & \textbf{36.7} & \textbf{38.2} & \textbf{41.0} \\ \hline
 \rowcolor{mygray}
\multicolumn{7}{c}{ORACLE: (Acc = 53.9)} \\ \hline
\multirow{2}{*}{SAT-ScanRefer~\cite{yang2021sat}} & \multicolumn{1}{c|}{baseline} & 36.7 & 40.6 & 43.7 & 50.3 & 37.0 \\
 & \multicolumn{1}{c|}{w. ours} & \textbf{51.6} & \textbf{44.5} & \textbf{47.6} & \textbf{52.5} & \textbf{50.7} \\ \hline
\end{tabular}%
\label{tab:refer_gt}
}
\end{minipage}
\vspace{-0.6cm}
\end{table*}

\vspace{-0.3cm}
\vspace{0.05in}\noindent\textbf{Evaluation Method.}
We benchmark different models of both 3D-VG and 3D-VQA. For 3D-VG with predicted proposals, we assess ScanRefer and MVT using the proposed \textit{ScanRefer-R}. For 3D-VG with Ground Truth (GT) proposals, we evaluate Referit3D and SAT~\cite{yang2021sat} on our \textit{NR3D-R}. In 3D-VQA, \textit{ScanQA-R} serves as the benchmark, including evaluations of ScanQA and 3D-LLM~\cite{hong20233d}. Except for SAT, all models are trained using their corresponding training sets. We explore both models for SAT from two training settings: training on original NR3D~\cite{achlioptas2020referit3d} and training on ScanRefer. 

\vspace{0.05in}\noindent\textbf{Metrics.}
For the 3D-VG task with predicted proposals, we use Acc@kIoU (k set to 0.25 and 0.5) to evaluate the accuracy, considering IoU thresholds of the box prediction. In 3D-VG tasks with ground truth proposals, we employ listening accuracy (Acc), where a model scores 1 for correctly selecting the target object from a list of candidates and 0 otherwise as defined in Referit3D~\cite{achlioptas2020referit3d}. For 3D-VQA, following ScanQA~\cite{azuma2022scanqa}, we report Exact Match@k (EM@k) for top-k prediction accuracy. Due to page limitations, other metrics, e.g.\ BLEU-1~\cite{papineni2002bleu}, CIDEr~\cite{vedantam2015cider} and their explanations, are provided in the supplementary file.

\vspace{-0.5cm}
\subsection{Main Results}
\begin{figure}[htbp]
  \centering
  \begin{subfigure}{0.21\linewidth}
    \includegraphics[height=2.0cm,keepaspectratio]{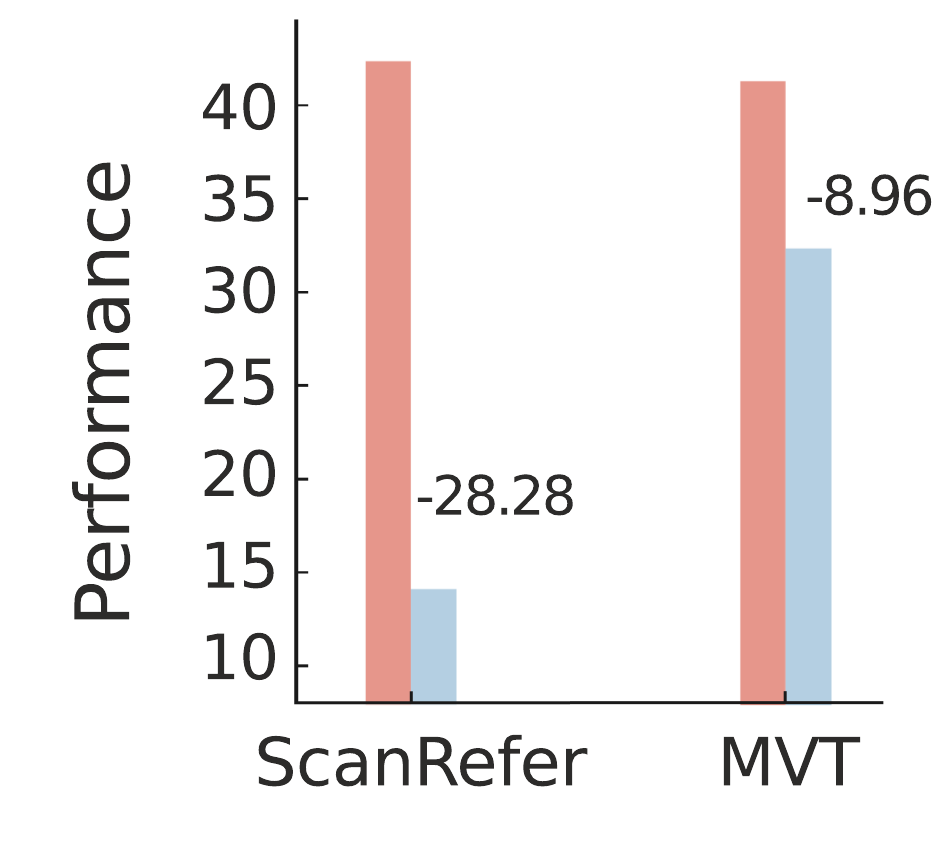}
    \caption{ScanRefer-pred}
    \label{fig:bar1}
  \end{subfigure}
  \begin{subfigure}{0.25\linewidth}
    \includegraphics[height=2.0cm,keepaspectratio]{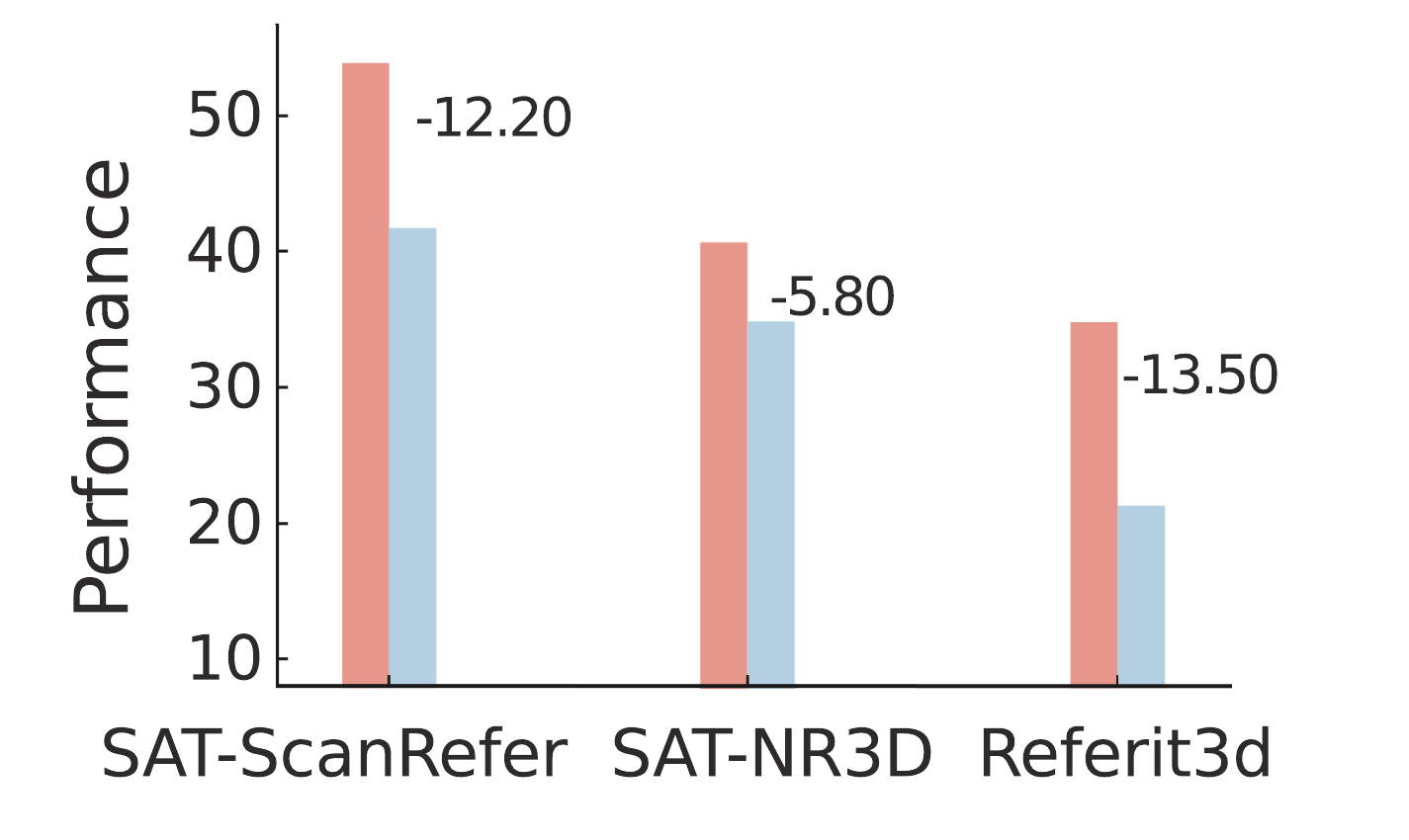}
    \caption{ScanRefer-GT}
    \label{fig:bar2}
  \end{subfigure}
  \begin{subfigure}{0.21\linewidth}
    \includegraphics[height=2.0cm,keepaspectratio]{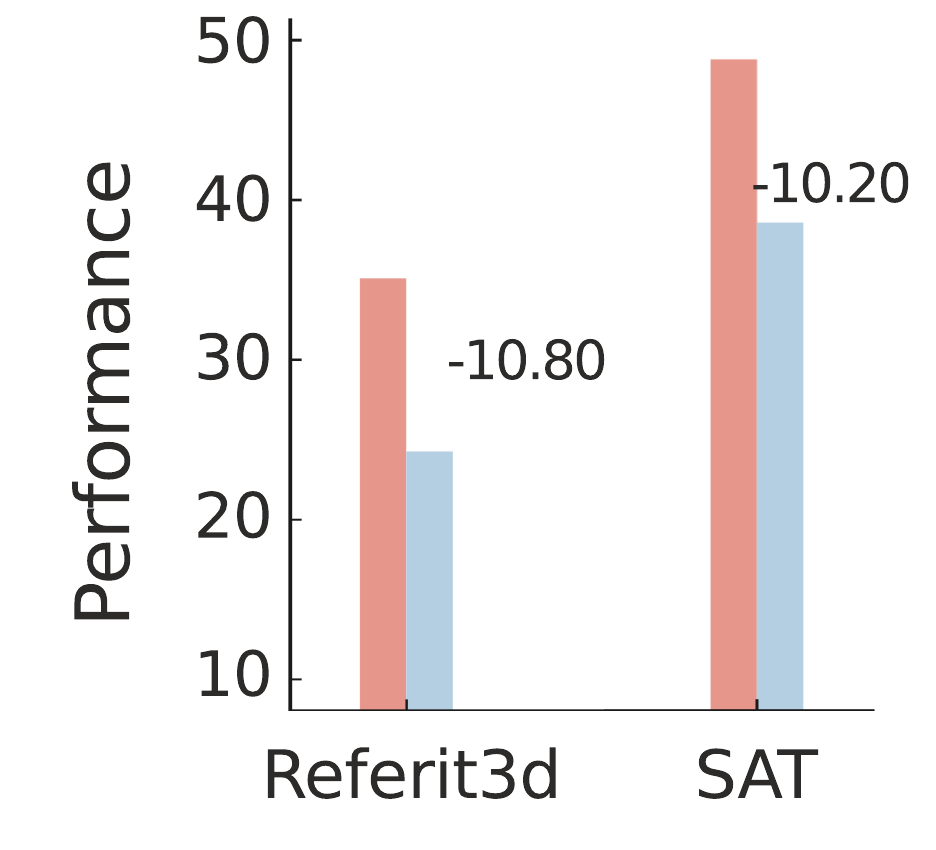}
    \caption{NR3D}
    \label{fig:bar3}
  \end{subfigure}
  \begin{subfigure}{0.25\linewidth}
    \includegraphics[height=2.0cm,keepaspectratio]{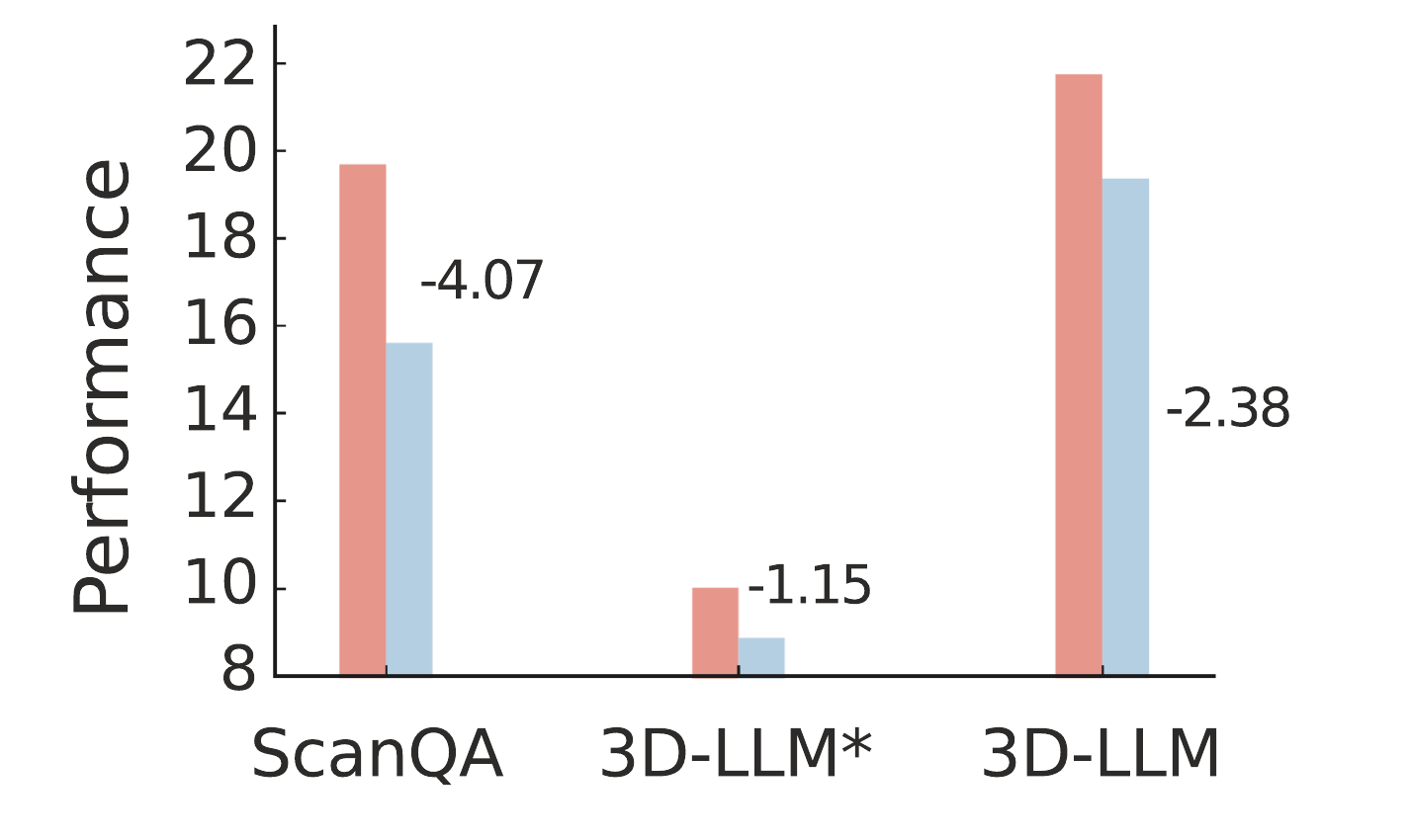}
    \caption{ScanQA}
    \label{fig:bar4}
  \end{subfigure}
  \begin{subfigure}{0.4\linewidth}
    \includegraphics[width=1\linewidth]{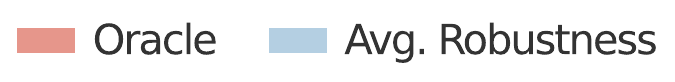}
    \label{fig:legend}
  \end{subfigure}
  \vspace{-20pt}
  \caption{Performance summary of existing models on our 3D Language Robustness benchmark. Listening accuracy (Acc) for NR3D~\cite{achlioptas2020referit3d} and ScanRefer-GT, Acc@kIoU for ScanRefer~\cite{chen2020scanrefer}, and EM@1 for ScanQA~\cite{azuma2022scanqa} are measured. We calculate average robustness by averaging its performance across all five language variant splits. It shows performance drops in 3D-QA and 3D-VG models, indicating a lack of robustness.}
  \vspace{-0.6cm}
  \label{fig:summary_res}

\end{figure}

We systematically evaluate various models for 3D-VG and 3D-VQA. The results are summarized in Fig.~\ref{fig:summary_res}. Oracle exhibits its performance on the unaltered test set, and we calculate average robustness by averaging its performance across all five language variant splits. It measures the overall performance degradation when the model encounters unfamiliar linguistic patterns (absent from the training set). We noted a serious performance drop in all models when confronted with language variants, indicating a lack of robustness in existing 3D-VL models.

\vspace{0.05in}\noindent\textbf{3D Visual Grounding.}
\cref{tab:scanrefer_box}(a) - \cref{tab:refer_gt}(c) present results for 3D Visual Grounding with Ground Truth (GT) and predicted bounding boxes on different datasets. Referit3D and SAT models (trained on different datasets) show significant performance decreases in tone and syntax splits, with Referit3D dropping by up to 18\% in simple syntax splits, (\cref{tab:refer_gt}(c)) indicating brittleness and overfitting to sentence patterns. This trend is consistent across datasets, including NR3D~\cite{achlioptas2020referit3d}, (\cref{tab:refer_gt}(b)) suggesting a bias towards the implicit structure in these datasets. Examining models with integrated point cloud detectors, like ScanRefer and MVT~\cite{huang2022multi}, we observe a similar pattern of sensitivity to language style changes, shown in \cref{tab:scanrefer_box}(a). ScanRefer shows a substantial performance decrease of 32.81\% in Acc@0.25 on tone splits, while MVT, using the more robust \textit{bert-based-uncased} text encoder, shows a smaller yet significant drop, particularly in syntax splits. These results highlight that existing 3D-VG models, regardless of whether they use detectors or not, are sensitive to variations in language styles.

\vspace{0.05in}\noindent\textbf{3D Visual Question Answering.}

\cref{tab:em_qa} outlines the 3D-VQA task results. Like in the 3D-VG task, the ScanQA model~\cite{azuma2022scanqa} shows performance degradation across all paraphrased style splits. The 3D-LLM model~\cite{hong20233d}, despite being pre-trained on a large language corpus, also experiences a reduction in performance in most splits but demonstrates greater robustness than other models. This resilience is attributed to its language model backbone's extensive training on a large corpus. Notably, in our modifier split, the un-fine-tuned 3D-LLM* variant surpasses the ORACLE in performance. This is likely because LLM-based models more effectively handle common language expressions, as seen in the modifier of nouns. Further discussions are shown in the supplementary file. Interestingly, 3D-LLM's fine-tuning on ScanQA faces a severe performance drop on the Tone split, suggesting that fine-tuning may cause catastrophic forgetting and lead to a biased feature space.

\begin{wraptable}{r}{5.8cm}
    \vspace{-1.0cm}
    \centering
    \hspace*{-0.4cm}
    \begin{small}
    \caption{Evaluating ScanQA and 3D-LLM, with/without our pre-alignment module on ScanQA. 3D-LLM* indicates the model before fine-tuning (FT) on ScanQA, while 3D-LLM shows post-FT results. Due to space limits, we only show EM@1 results; additional metrics are in the Suppl.. Best results are in \textbf{bold}.}
    \scalebox{0.7}{
        \begin{tabular}{cccccc}
            \hline
            \multicolumn{1}{c|}{Method} & Syntax & Voice & Modifier & Accent & Tone \\ \hline
            \rowcolor{mygray}
            \multicolumn{6}{c}{ORACLE: (EM@1 = 19.69)} \\ \hline
            \multicolumn{1}{c|}{ScanQA~\cite{azuma2022scanqa}} & 14.98 & 17.12 & 16.01 & 14.04 & 15.92 \\
            \multicolumn{1}{c|}{w. ours} & \textbf{19.26} & \textbf{18.58} & \textbf{18.32} & \textbf{18.41} & \textbf{18.49} \\ \hline
            \rowcolor{mygray}
            \multicolumn{6}{c}{ORACLE: (EM@1 = 10.02)} \\ \hline
            \multicolumn{1}{c|}{3D-LLM*~\cite{hong20233d}} & 8.99 & 9.50 & 10.27 & 8.13 & 7.45 \\
            \multicolumn{1}{c|}{w. ours} & \textbf{10.19} & \textbf{9.59} & 9.93 & \textbf{9.25} & \textbf{9.25} \\ \hline
            \rowcolor{mygray}
            \multicolumn{6}{c}{ORACLE: (EM@1 = 21.75)} \\ \hline
            \multicolumn{1}{c|}{3D-LLM~\cite{hong20233d}} & 21.06 & 19.61 & 19.86 & 19.52 & 16.78 \\
            \multicolumn{1}{c|}{w. ours} & \textbf{21.15} & 19.52 & \textbf{20.80} & \textbf{19.95} & \textbf{20.80} \\ \hline
        \end{tabular}%
        \label{tab:em_qa}
    }
    \end{small}
    
    \vspace{-0.7cm}
    
\end{wraptable}

\section{Analysis and Improved Model}
The systematic evaluation results indicate a lack of robustness in existing 3D-VL models. We analyze the reason behind this issue through an in-depth investigation. (Sec.~\ref{sec5:reason}) Next, we developed a simple yet effective plug-and-play method. (Sec.~\ref{sec5:ours}) This approach can be applied to any pre-trained model, enhancing its robustness to diverse natural language characteristics without the need for re-training or additional data augmentation. Finally, we discuss data augmentation, a technique commonly used to enhance model performance and robustness. (Sec.~\ref{sec5:aug})

\subsection{Why do the 3D-VL models fail?}  
\label{sec5:reason}

Our systematic evaluation identifies the lack of robustness is consistently and commonly observed across various models and tasks. Suggesting existing models fail at aligning 3D modality with text. We further conduct an in-depth analysis to figure out ``Why do 3D-VL models fail?''.

\vspace{0.05in}\noindent\textbf{The Fragility of Fusion Space.}
We notice that in \cref{tab:scanrefer_box}(a), 
though MVT~\cite{huang2022multi} is equipped with a pre-trained BERT~\cite{Devlin2019BERTPO}, a transformer-based language model, it still faces performance dropping issues. Therefore, we hypothesize the real problem is the feature fusion module used to fuse point cloud features generated by the vision encoder ~\cite{qi2017pointnet,qi2017pointnet++, qi2019deep} and text features generated by the text encoder.

\begin{figure}[htbp]
  \centering
  \begin{subfigure}{0.4\linewidth}
    \includegraphics[width=1\linewidth]{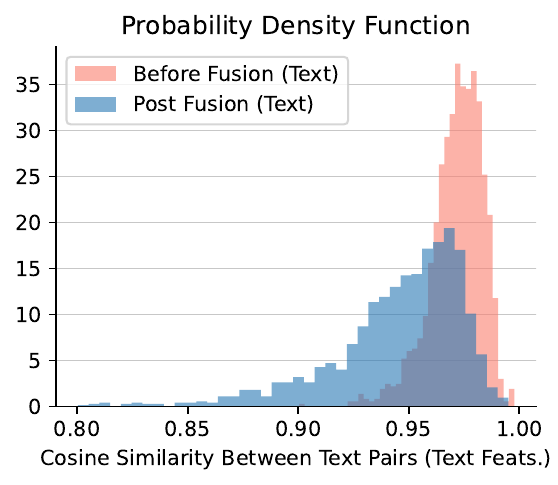}
    \caption{Text feature similarity.}
    \label{fig:pdf_text}
  \end{subfigure}
  \begin{subfigure}{0.4\linewidth}
    \includegraphics[width=1\linewidth]{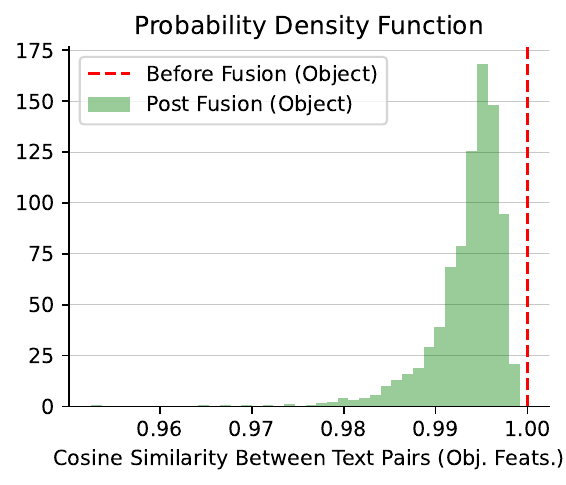}
    \caption{Object feature similarity.}
    \label{fig:pdf_obj}
  \end{subfigure}
  \vspace{-5pt}
  \caption{Probability Density Functions (PDFs) of cosine similarity between text pairs of original Sentences and their corresponding syntax variants.}
  \vspace{-0.8cm}
  \label{fig:pdf_analysis}
\end{figure}

To study this, we use the syntax variants from our \textit{ScanRefer-R} as an example. We measure the sentence vector cosine similarity between the original sentence and the syntax variants before the fusion module (right after the BERT, shown in Fig.~\ref{fig:archs}(a)) and after passing the fusion module. 
We analyze all failure cases in this split by calculating the similarity between sentence embedding features or object features of the syntax variants and their corresponding original sentences. Subsequently, we construct the probability density function (PDF), as illustrated in Fig.~\ref{fig:pdf_analysis}. The PDF for cosine similarity between text pairs reveals distinct patterns before and after the fusion process. For text features, distribution before fusion (red distribution in Fig.~\ref{fig:pdf_analysis}(a)) is notably skewed towards higher similarity values, suggesting that the original sentences and their syntax variations (not seen in the training set) are closely aligned. This indicates that the text encoder (e.g., BERT) is robust enough to handle variations well.

In contrast, the cosine similarity distribution of text features post-fusion (blue distribution in Fig.~\ref{fig:pdf_analysis}(a)) is more widespread, with a significant portion extending towards lower similarity values. This suggests that the fusion process negatively affects the integrity of text features. The robustness of the fusion module is weak towards sentence variations. It performs poorly when presented with syntax-altered sentences that are absent from the training data. In other words, the fusion module is biased towards the training dataset, rather than genuinely understanding the semantics of natural language. 

Moreover, Fig.~\ref{fig:pdf_analysis}(b) depicts the distribution of object features, calculated analogously to the text similarity previously discussed. Given that the input visual scene remains constant for both the sentences in the syntax split and the original sentences, the expected similarity value should be exactly 1. Nonetheless, the green distribution in the figure shifts towards the left, indicating lower similarity values. This shift mirrors the trend observed in the text feature distribution, implying a negative impact on the object features as well. This further substantiates the fragility of the fusion feature space and the bias problem directed toward the training set, rather than genuinely understanding the semantics of natural language. Besides, it is worth mentioning that the absolute value of similarity \textit{does not hold physical meaning} in high-dimension space, therefore we \textit{focus on the relative comparison results}.

\vspace{0.05in}\noindent\textbf{Low Diversity of Existing Datasets.}
We further identify that the major cause behind the above phenomenon is rooted in the low diversity of the original dataset and its large domain gap to the human daily language. It hinders the model from learning more general information. For example, we study the syntax diversity using the ScanRefer~\cite{chen2020scanrefer} dataset. Specifically, we propose to use a vectorized syntax tree to represent a sentence, then we conduct PCA analysis~\cite{mackiewicz1993principal} and plot the density map. The darker region shows more sentences are being projected to this place. We can tell from the density map that the origin 3D-VL dataset ScanRefer has a simpler syntax pattern (the more compact darker region in Fig.~\ref{fig:densit}(a)) while the other dataset shows more spreading high-density regions. It indicates that the existing dataset lacks enough diversity to reflect real-world language characteristics properly. Moreover, given such a dataset to facilitate model training, the model can easily fit the simple implicit pattern without obtaining a generic understanding of language features. It is worth mentioning that our proposed dataset (Fig.~\ref{fig:densit}(b)) shares a similar pattern with the open assistant~\cite{kopf2023openassistant} dataset (Fig.~\ref{fig:densit}(d)), which is a dataset that has a closer domain gap with daily language and enough diversity, indicating the high quality of our dataset. We further prove that increasing the dataset's diversity can reduce the robustness problem to a certain extent, as detailed in the discussion of data augmentation in Sec.~\ref{sec5:aug}.

\vspace{-0.7cm}

\vspace{-0.3cm}
\subsection{Plug and Play Pre-Alignment Module}
\label{sec5:ours}
\vspace{-0.2cm}

\vspace{0.05in}\noindent\textbf{Pre-Alignment Module.}
We have identified that the fusion module is sensitive to the textual input and shows overfitting to the language style pattern of training data. Therefore, we propose a training-free pre-alignment method to first convert the sentence pattern into the style that the trained model prefers. More specifically, we design an LLM-based parser, which maps sentences in any style into the models' preferred style. Since we have a well-trained model, we naturally assume that we can assess the training data. We propose to use LLM to conduct the style transfer, mapping sentences in any style into the format that the model is good at while maintaining the same meaning. We design a structured prompt containing \uline{``Generic Rules''} and \uline{``In-Context Examples''}: the genetic rule contains format instructions for better post-processing, a simple sentence \textit{``You should not change the meaning of the input sentence...''} to encourage the model to preserve most information. Since we can see what kind of data that model is trained from, we just need to give three to six in-context examples indicating the pattern we are dealing with and the style model preferred. Empirical evidence suggests that for modern, powerful large language models, as few as 3 to 6 samples are sufficient to capture the desired style effectively. The pre-aligned sentence is directly fed into the same model for prediction. This process is called pre-alignment without the need for retraining. It also does not need large-scale additional annotation on different variants. 

\vspace{0.05in}\noindent\textbf{Performance Benefit of Our Pre-Alignment Module.}

When used in conjunction with existing models, our pre-alignment module achieves significant performance enhancements across various datasets, as demonstrated in \cref{tab:scanrefer_box} to \cref{tab:em_qa}. In particular, ScanRefer and MVT models exhibit substantial improvements across all language variants on \textit{ScanRefer-R} with our module. For example, ScanRefer's Acc@0.25 increases by over 16\% in the tone split. Similar improvements are observed in \textit{NR3D-R} and 3D-VQA models. However, 3D-LLM~\cite{hong20233d} without fine-tuning does not benefit from our module because it effectively handles language expressions since it is a large-scale pre-trained language model. 
\cref{tab:aug_table} further shows the performance of our proposed method with SAT (a grounding model). Our pre-alignment module recovered performance on all five splits compared with this baseline. Notably, our method can recover up to 12\% accuracy on tone split without training. This indicates the effectiveness of our proposed module. More discussion is provided in the supplementary file.

\vspace{-0.2cm}
  \subsection{Discussion on Data Augmentation}
\label{sec5:aug}
\vspace{-0.1cm}

Here, we investigate whether data augmentation can address model robustness. We experimented with the NR3D~\cite{achlioptas2020referit3d} dataset in two scenarios, with the result shown in \cref{tab:aug_table}. The first, ``SAT w. aug - 40k'', involved training on a mixed-style dataset, size-matched to the original NR3D, resulting in 40,000 samples balanced on five variants. The second, ``SAT w. aug - 80k'', doubled the dataset to 80,000 samples by merging training data variants. We discuss more details in the supplementary. \cref{tab:aug_table} shows our method surpasses data-augmented models of equal size and rivals those trained on double the data. This highlights our method's effectiveness, especially considering the impracticality and high cost of obtaining exhaustive annotations for augmentation. Which also demonstrates the significant data demand for augmentation.

\begin{wraptable}{r}{5.8cm}
    \vspace{-1.1cm}
    \centering
    \hspace*{-0.4cm}
    \begin{small}
    \caption{Results of data augmentation. Comparing results of SAT training with or without NR3D-R variants as extra data on NR3D, measured in listening accuracy. The best results are in \textbf{bold}.}
    \scalebox{0.7}{
        \begin{tabular}{c|ccccc}
        \hline
            Backbone & Syntax & Voice & Modifier & Accent & Tone \\ \hline
            SAT~\cite{yang2021sat} & 43.9 & 34.2 & 38.9 & 41.2 & 34.9 \\
            SAT w. aug - 40k & 43.1 & 40.8 & 40.5 & 41.0 & 43.5 \\
            SAT w. aug - 80k & \textbf{45.7} & 42.3 & \textbf{44.5} & 42.5 & 45.9 \\
            SAT w.ours & 45.1 & \textbf{44.1} & 41.6 & \textbf{44.6} & \textbf{46.9} \\ \hline
        \end{tabular}
        \label{tab:aug_table}
    }
    \end{small}
    \vspace{-0.8cm}
    
\end{wraptable}

Both data augmentation and our method hold potential.\underline{(i)} We believe augmentation could improve performance, but it demands large and diverse high-quality datasets (\cref{tab:aug_table}). However, as human language exhibits over 23 styles, it is difficult to cover all for augmentation. \underline{(ii)} Our method, requiring no training, offers immediate plug-and-play benefits to existing models and is ready for future boosts with more advanced LLMs. \underline{(iii)} Our approach could offer diverse data for augmentation. 
\vspace{-0.4cm}
\section{Conclusion}
\vspace{-0.2cm}
We present a novel task and benchmark dataset to study language robustness in 3D vision-language models. Our dataset, designed based on linguistic theories, includes diverse language variants. Through comprehensive evaluations, we identify significant fragility in existing 3D-VL models when encountering real-world language variations. We analyze the underlying reasons and propose a simple yet effective solution without requiring additional training. This work aims to facilitate future research in improving model robustness, crucial for deploying 3D-VL models in real-world applications.

\bibliographystyle{splncs04}
\bibliography{main}

\clearpage
\appendix

\section{More Examples}
\label{sec:examples}
\begin{figure*}[htbp]
  \centering
   \includegraphics[width=0.9\textwidth]{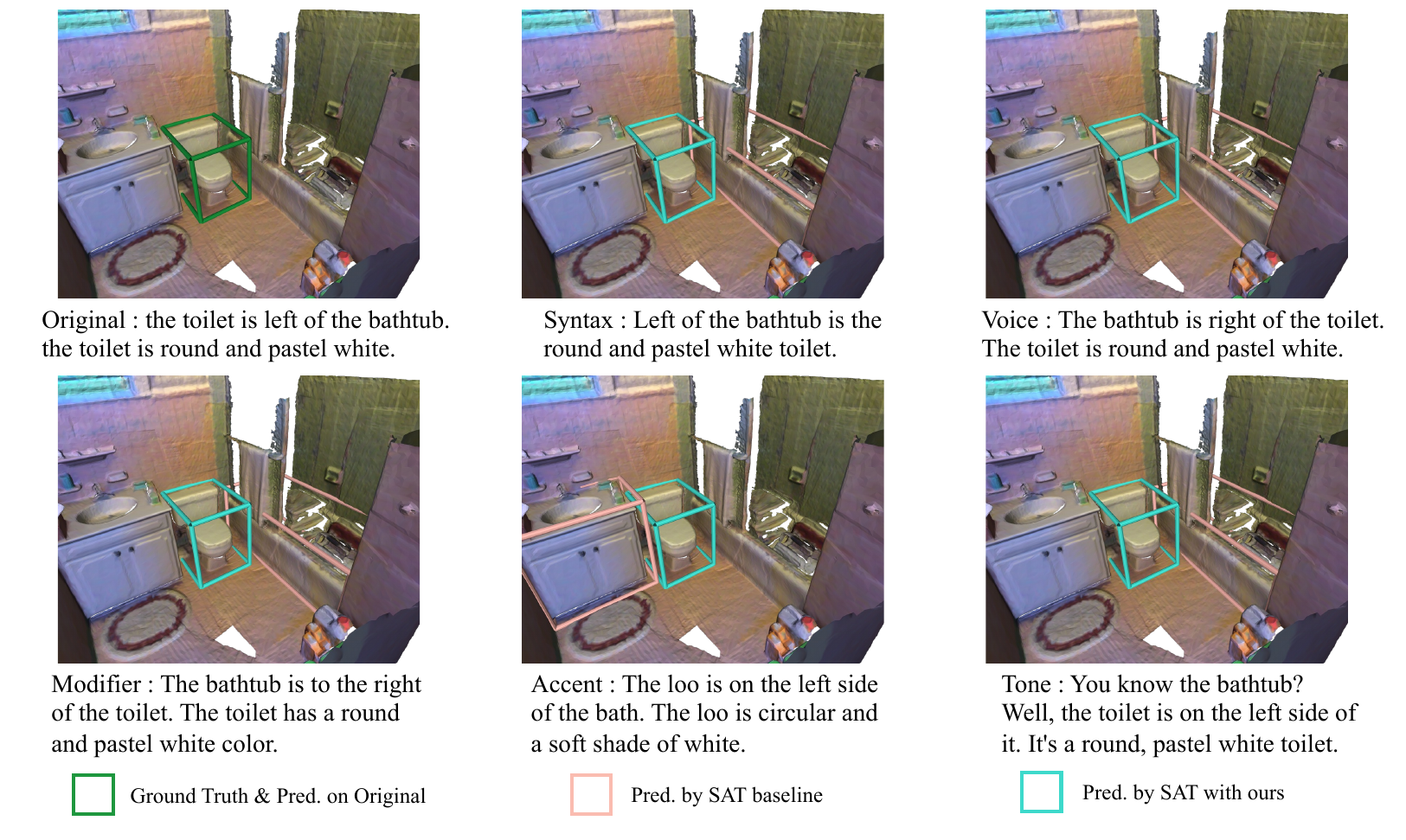}

   \caption{Predictions by SAT~\cite{yang2021sat} (a 3D Visual Grounding Model) on ScanRefer's sentence and five variants, compared with ``SAT with Ours'' using our pre-alignment module. Ground truth prediction is green-highlighted and predictions from plain SAT are pink-highlighted. The figure shows SAT's difficulty with language variants, while our plug-and-play module aids in accurate predictions.
   }
   \label{fig:example_ours_1}
\end{figure*}

\begin{figure*}[htbp]
  \centering
   \includegraphics[width=0.9\textwidth]{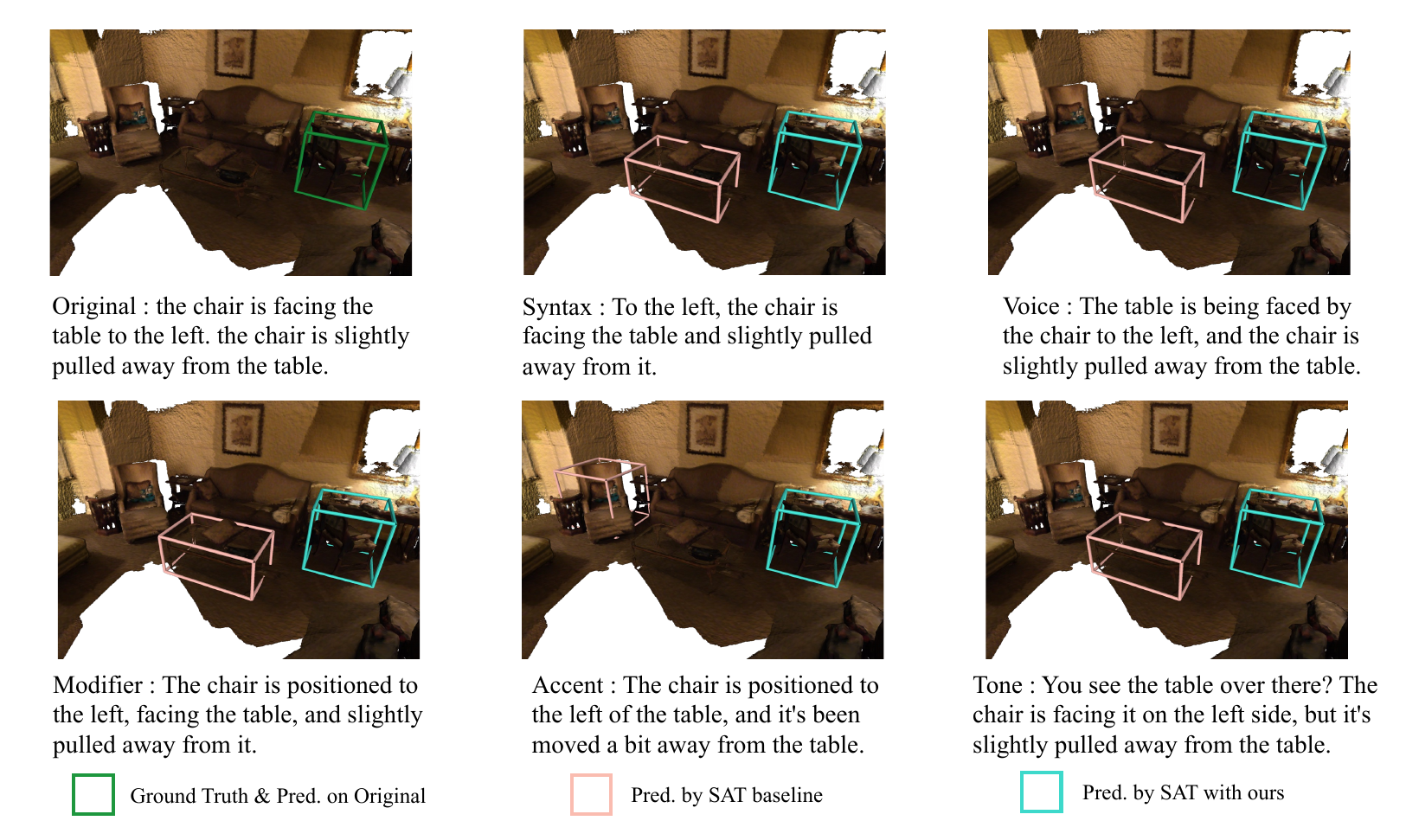}

   \caption{Predictions by SAT~\cite{yang2021sat} (a 3D Visual Grounding Model) on ScanRefer's sentence and five variants, compared with ``SAT with Ours'' using our pre-alignment module. Ground truth prediction is green-highlighted and predictions from plain SAT are pink-highlighted. The figure shows SAT's difficulty with language variants, while our plug-and-play module aids in accurate predictions.}
   \label{fig:example_ours_2}
\end{figure*}

This section provides additional visual examples to further illustrate the language robustness challenge discussed in the main paper. Examples are presented in Fig.~\ref{fig:example1} to Fig.~\ref{fig:example5}. We employ SAT~\cite{yang2021sat}, a 3D Visual Grounding model trained on the ScanRefer dataset~\cite{chen2020scanrefer}, to test its effectiveness in handling subtle linguistic variations. Specifically, we analyze its response to five sentence variants that are semantically identical to the original but phrased differently, reflecting natural variations in human communication. These examples underscore a key limitation: the model struggles with minor rephrasings, such as sentence inversions, that do not change the underlying meaning, highlighting a critical challenge in current natural language processing capabilities. Moreover, we visualize the predictions, shown in Fig.~\ref{fig:example_ours_1} and Fig.~\ref{fig:example_ours_2}, made by plain SAT and SAT enhanced with our pre-alignment module to demonstrate the effectiveness of our approach.

\section{Datasets}\label{sec:ds}
In this section, we first present basic statistics and semantic similarity analysis for two additional splits within our \textbf{3D} \textbf{L}anguage \textbf{R}obustness (\textbf{3D-LR}) dataset: \textit{NR3D-R} for 3D Visual Grounding (3D-VG) and \textit{ScanQA-R} for 3D Visual Question Answering (3D-VQA). We then delve into a detailed discussion about the variations in language usage within these datasets. Additionally, we outline the methodology employed in leveraging large language models (LLMs) to construct the \textbf{3D-LR} dataset. Lastly, we discuss the diversity of these datasets.
\subsection{Basic Statistics}
\cref{tab:more_stas} detail the basic statistics of \textit{ScanQA-R} and \textit{NR3D-R} respectively. They compare original sentences with their rephrased counterparts, focusing on characteristics such as word count, noun usage, and sentence complexity. These statistics are akin to those observed for \textit{ScanRefer-R}, as discussed in Section 3.2 of the main paper. This similarity indicates that our strict rephrasing rules effectively preserve the original meaning and core object nouns across different dataset splits. The consistent number of unique words across these splits corroborates this observation. For the modifier, accent, and tone variation, which aims to mimic conversational style, we incorporated additional verbal phrases and modifiers into the original sentences. This modification led to an increase in the unique words, total words, and average sentence length. Beside, we also use both neural semantic metrics and tradition metrics to ensure our dataset preserve the original meaning. The results are shown in \cref{tab:more_sem_sim}. We can observe that all our paraphrased dataset has an edit distance to the original sentence lower than 1, which means they are only minorly different from the original sentence. Moreover, the high cosine similarity in BERT and Glove suggests that, from a neural network model's perspective, they almost mean the same.

\begin{table*}[t]
\centering
\caption{
More basic statistics. We count the unique words, total words, and the average sentence length on our paraphrased dataset, named \textit{ScanQA-R}, in (a) and \textit{NR3D-R} in (b).
}
\vspace{-0.2cm}
\label{tab:more_stas}

\begin{minipage}[b]{0.47\linewidth}
\centering
\caption*{\textbf{(a)} Basic statistics of different data splits using \textit{ScanQA-R}
as an example.}

\vspace{-0.3cm}
\centering
\resizebox{\columnwidth}{!}{%
\begin{tabular}{c|ccc}
\hline
\multicolumn{1}{l|}{} & Unique words & Total words & Avg. description length \\ \hline
Original              & 565          & 10257       & 8.78                    \\
Syntax                & 556          & 10731       & 9.19                    \\
Voice                 & 566          & 10665       & 9.13                    \\
Modifier              & 611          & 11548       & 9.89                    \\
Accent                & 679          & 14085       & 12.06                   \\
Tone                  & 606          & 14385       & 12.32                   \\ \hline
\end{tabular}%
}
\end{minipage}
\hfill
\begin{minipage}[b]{0.48\linewidth}
\centering
\caption*{\textbf{(b)} 
Basic statistics of different data splits using \textit{NR3D-R}
as an example.
}
\vspace{-0.3cm}
\centering
\resizebox{\columnwidth}{!}{%
\begin{tabular}{c|ccc}
\hline
\multicolumn{1}{l|}{} & Unique words & Total words & Avg. description length \\ \hline
Original              & 909          & 19475       & 10.41                   \\
Syntax                & 907          & 20769       & 11.11                   \\
Voice                 & 928          & 21785       & 11.65                   \\
Modifier              & 1047         & 22616       & 12.09                   \\
Accent                & 1002         & 22857       & 12.22                   \\
Tone                  & 873          & 30702       & 16.42                   \\ \hline
\end{tabular}%
}
\end{minipage}
\vspace{-0.6cm}
\end{table*}

\subsection{More Discussion of Language Variants}

\begin{figure*}[htbp]
  \centering
   \includegraphics[width=0.9\textwidth]{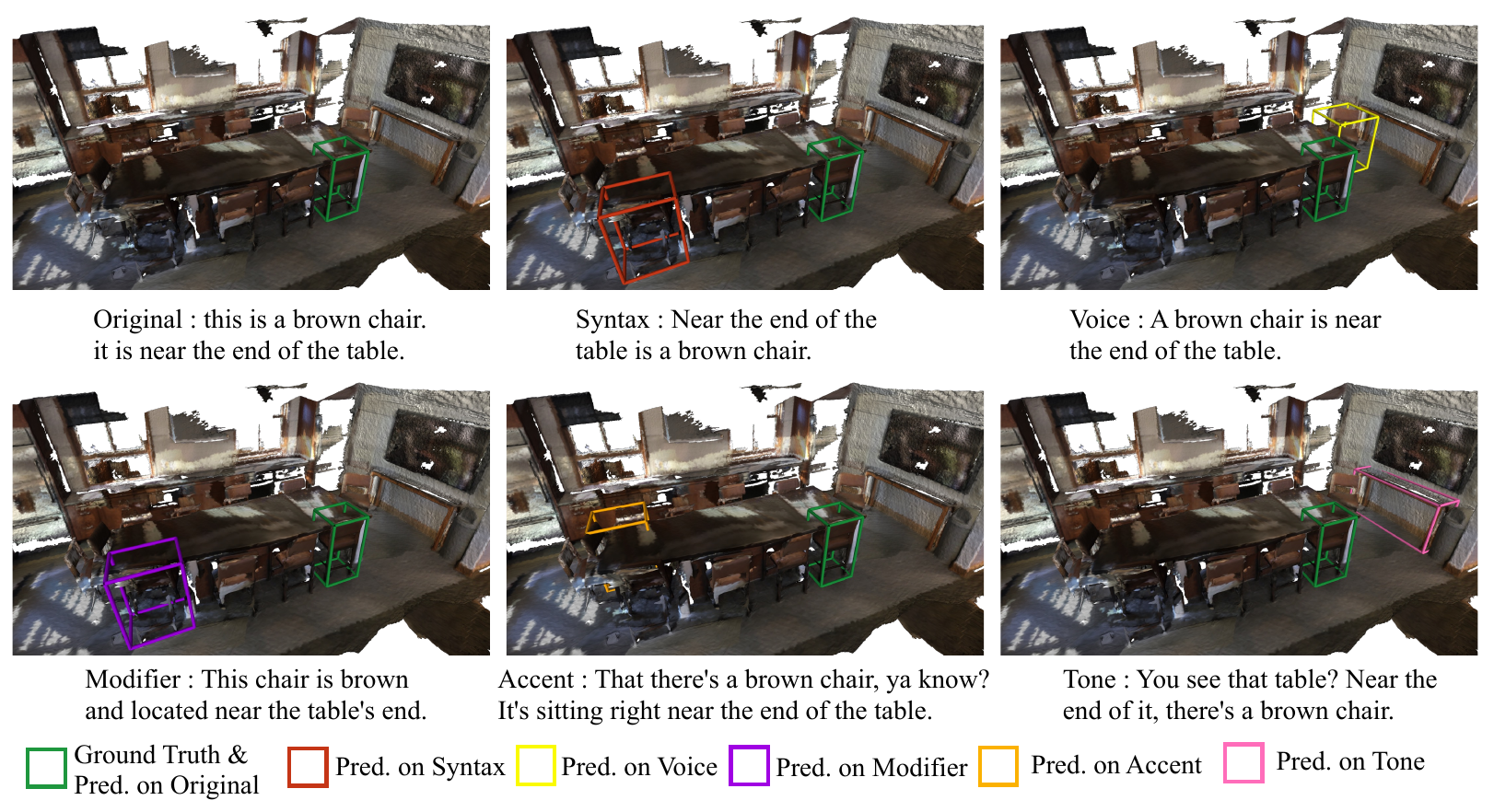}

   \caption{Prediction made by SAT~\cite{yang2021sat} (a 3D Visual Grounding Model) on ScanRefer's original sentence and five modified variants. The ground truth and the SAT's prediction for the original sentence are highlighted in green. Differently colored boxes indicate the SAT's incorrect predictions on the sentence variants.
   }
   \label{fig:example1}
\end{figure*}

\begin{figure*}[htbp]
  \centering
   \includegraphics[width=0.9\textwidth]{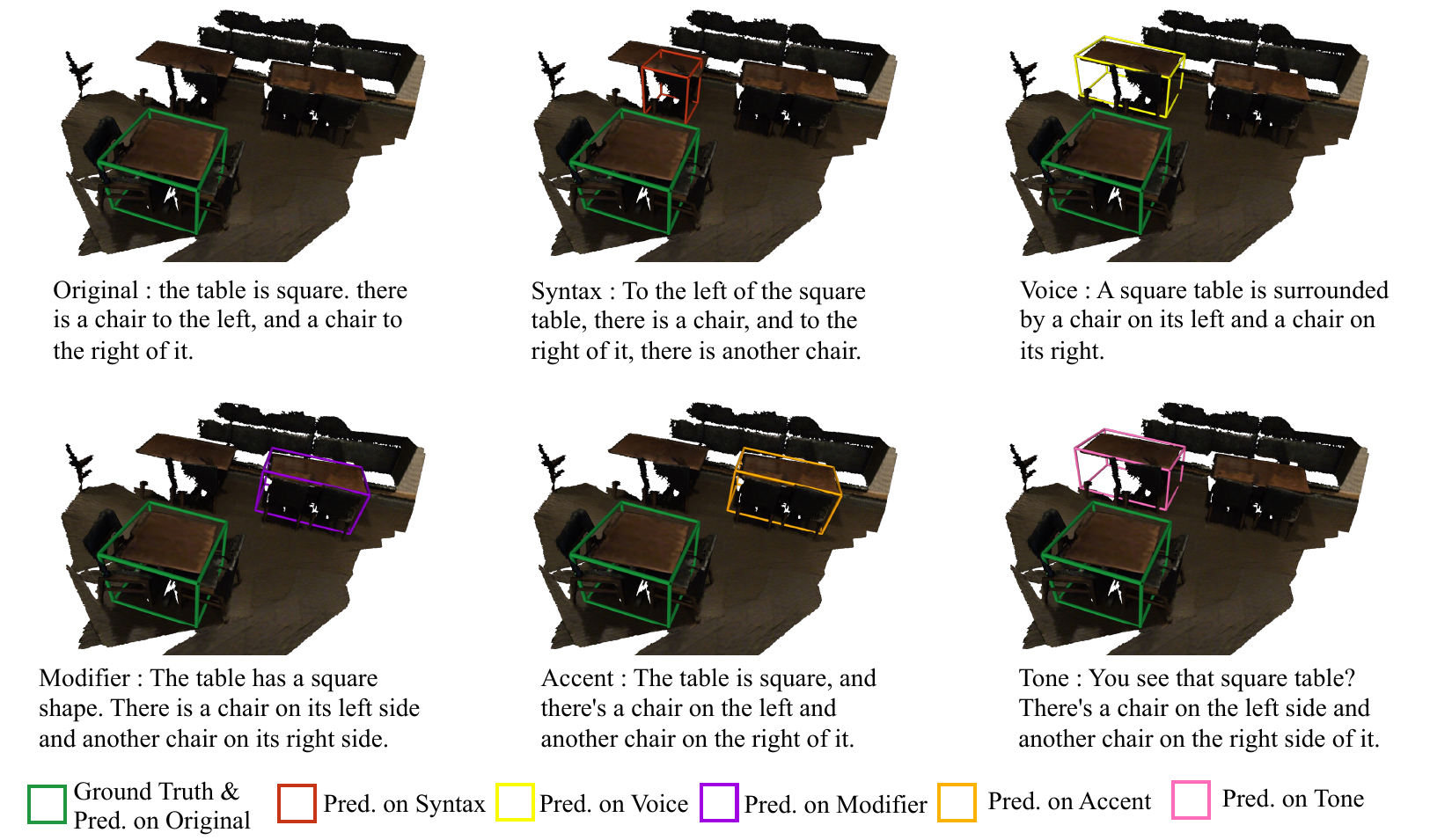}

   \caption{Prediction made by SAT~\cite{yang2021sat} (a 3D Visual Grounding Model) on ScanRefer's original sentence and five modified variants. The ground truth and the SAT's prediction for the original sentence are highlighted in green. Differently colored boxes indicate the SAT's incorrect predictions on the sentence variants.
   }
   \label{fig:example2}
\end{figure*}

\begin{figure*}[htbp]
  \centering
   \includegraphics[width=0.9\textwidth]{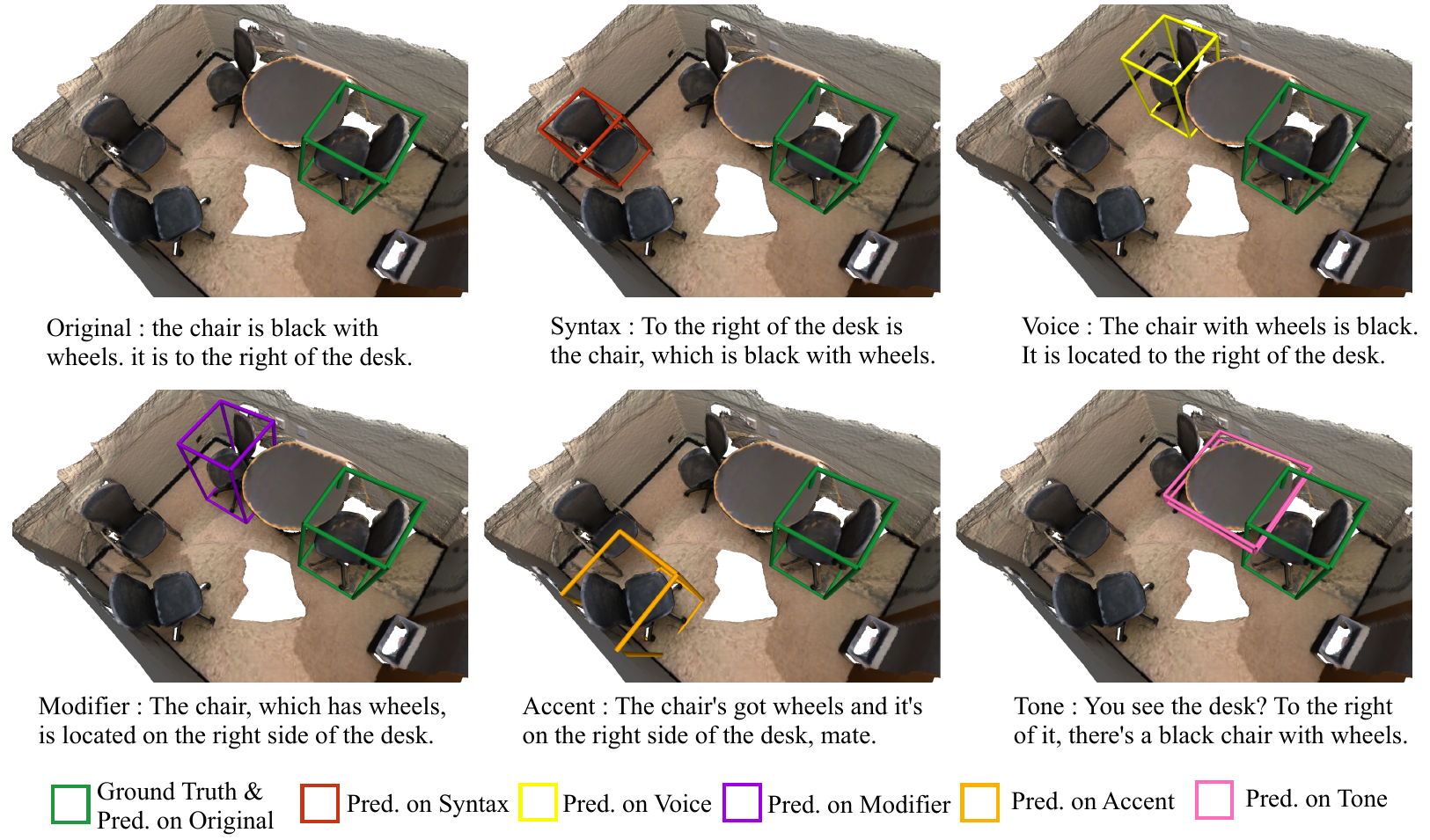}

   \caption{Prediction made by SAT~\cite{yang2021sat} (a 3D Visual Grounding Model) on ScanRefer's original sentence and five modified variants. The ground truth and the SAT's prediction for the original sentence are highlighted in green. Differently colored boxes indicate the SAT's incorrect predictions on the sentence variants.
   }
   \label{fig:example3}
\end{figure*}

\begin{figure*}[htbp]
  \centering
   \includegraphics[width=0.9\textwidth]{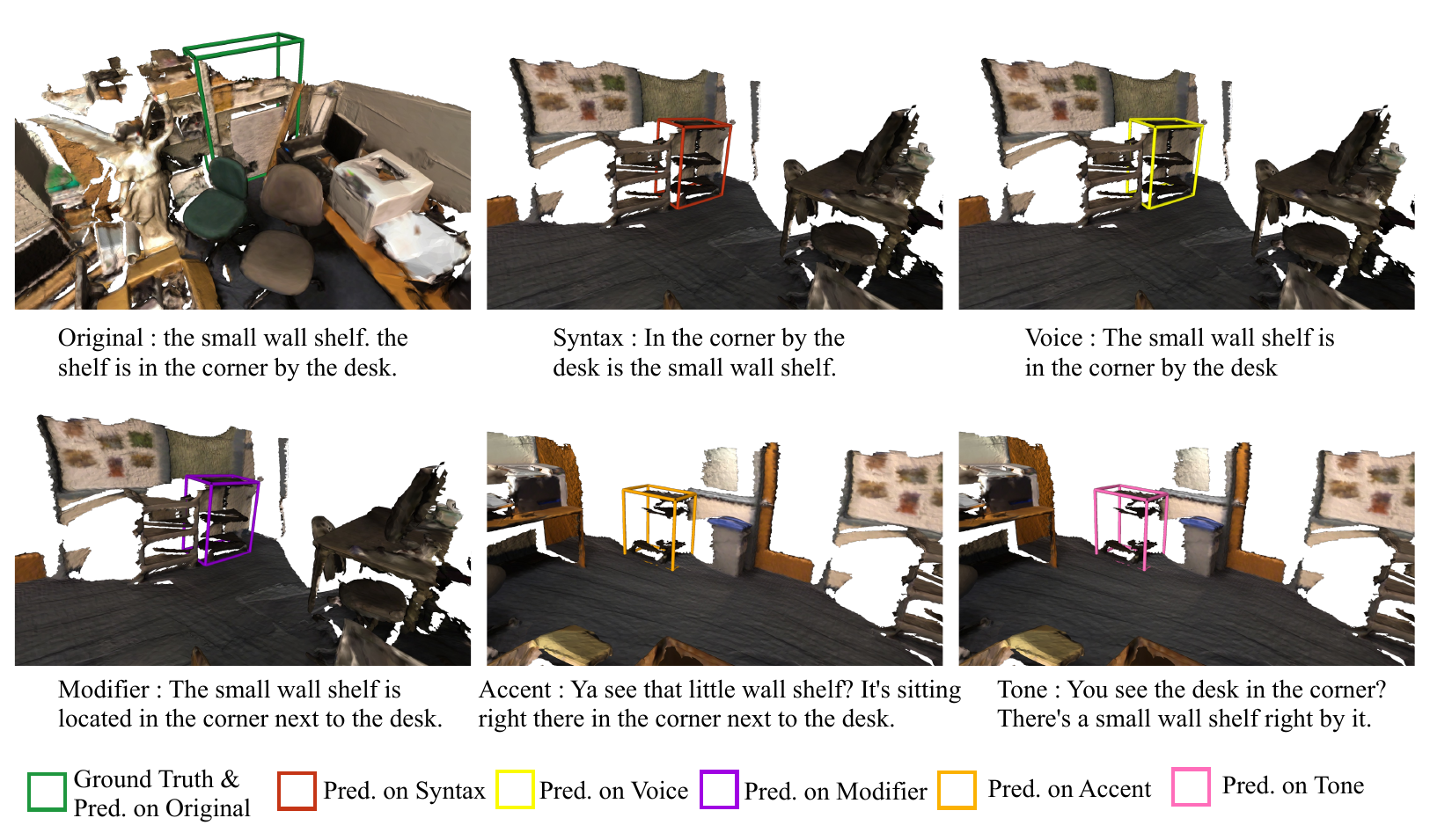}

   \caption{Prediction made by SAT~\cite{yang2021sat} (a 3D Visual Grounding Model) on ScanRefer's original sentence and five modified variants. The ground truth and the SAT's prediction for the original sentence are highlighted in green. Differently colored boxes indicate the SAT's incorrect predictions on the sentence variants.
   }
   \label{fig:example4}
\end{figure*}

\begin{figure*}[htbp]
  \centering
   \includegraphics[width=0.9\textwidth]{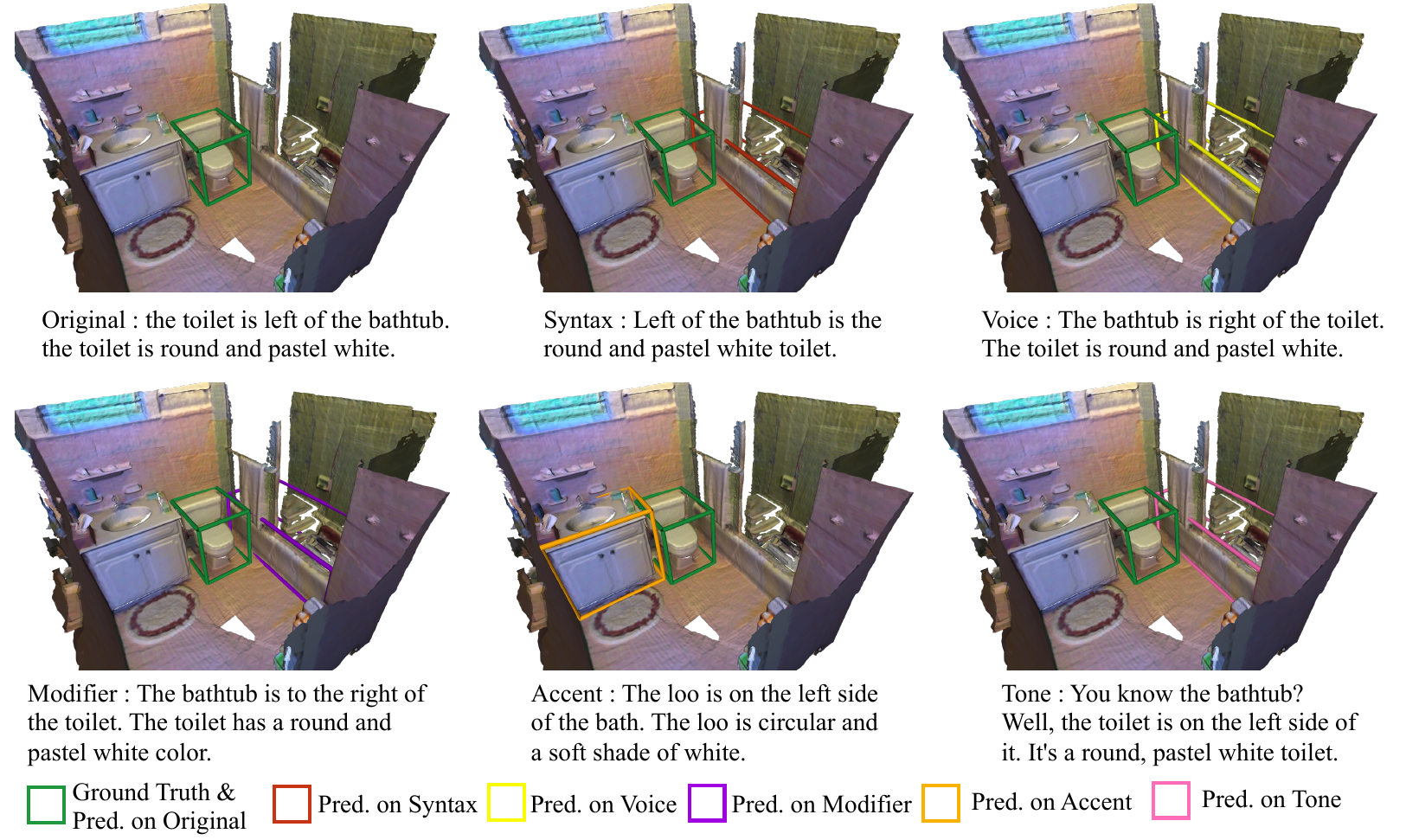}

   \caption{Prediction made by SAT~\cite{yang2021sat} (a 3D Visual Grounding Model) on ScanRefer's original sentence and five modified variants. The ground truth and the SAT's prediction for the original sentence are highlighted in green. Differently colored boxes indicate the SAT's incorrect predictions on the sentence variants.
   }
   \label{fig:example5}
\end{figure*}

We provide several examples of sentence variants representing different aspects of human language as illustrated in the captions below the corresponding images from Fig.~\ref{fig:example1} to Fig.~\ref{fig:example5}. These examples showcase five distinct groups of sentences derived from the original sentences in the ScanRefer dataset, highlighting variations in syntax, voice, modifier, accent, and tone.

Humans employ many language styles to convey their intentions, meanings, and emotions and emphasize certain aspects~\cite{barber_beal_shaw_2009}. These language styles allow for subjective expressions and the inclusion of emotions, enriching the communication experience.

\textbf{Syntax}, for instance, plays a crucial role in human language by providing various word or phrase orders to create different sentence structures. Inverse sentences, such as posing a question before providing the answer or lifting the part that one wants to highlight to the start of the sentence, are common in daily conversations to engage the listener and evoke curiosity or anticipation.

\textbf{Voice}, on the other hand, enables individuals to transform sentences from active to passive voice or vice versa. This intentional use of voice enables slight changes in emphasis, highlighting different subjects or actions in a sentence, and can evoke various emotional responses from the listener.

\textbf{Modifiers}, such as adjectives and adverbs, are extensively used by humans to enhance the details and depth of their expressions. By incorporating a wide variety of modifiers, individuals can add subjective elements, convey emotions, and paint vivid pictures in the minds of their audience.

\textbf{Accent} reflects the distinct linguistic habits of English speakers from different regions, contributing to the diversity and richness of language. It allows individuals to express their cultural identity and provides a unique flavor to their communication, introducing subjective nuances and regional colloquialisms. 

Lastly, \textbf{tone} encompasses the attitudes and emotions conveyed in a sentence. For example, by using questions, employing irony, or other techniques, individuals can add layers of meaning and emphasize certain aspects, evoking specific emotional responses from their audience.

\subsection{Detailed Prompt for Dataset Constructions}
We employ structured prompting and chain of thought~\cite{wei2022chain} techniques to create language variants. Please note that the ``style requiremen'' slot specifies the desired variant type in strict rules and the ``sentence'' slot should be filled with the original sentence for rewriting. Our process ensures high-quality rewrites while adhering to these parameters. Fig.~\ref{fig:prompt} shows the overall structure of our prompt.

\begin{table*}[t]
\centering
\caption{
More semantic similarity statistics. We measure the the cosine similarity between our \textit{ScanQA-R} (a) and \textit{NR3D-R} (b) with the original data using BERT embedding, Glove embedding. We also compare the Edit Distance (ED). This suggests that our data effectively retains the original's meaning.
}
\vspace{-0.2cm}
\label{tab:more_sem_sim}

\begin{minipage}[b]{0.48\linewidth}
\centering
\caption*{\textbf{(a)} 
Semantic similarity analysis of our \textit{ScanQA-R}.
}
\vspace{-0.3cm}
\centering
\scalebox{0.75}{
\begin{tabular}{cccc}
      \toprule
      & BERT Sim. ($\uparrow$) & Glove Sim. ($\uparrow$) & ED ($\downarrow$) \\
      \midrule
      Syntax & 0.94 & 0.98 & 0.40 \\
      Voice & 0.96 & 0.98 & 0.72 \\
      Modifier & 0.94 & 0.93 & 0.65 \\
      Accent & 0.84 & 0.85 & 0.53 \\
      Tone & 0.87 & 0.92 & 0.48 \\
      \bottomrule
    \end{tabular}%
    \label{tab:more_semantic_sim_2}
}
\end{minipage}
\hfill
\begin{minipage}[b]{0.47\linewidth}
\centering
\caption*{\textbf{(b)} Semantic similarity analysis of our \textit{NR3D-R}.}
\vspace{-0.3cm}
\centering
\scalebox{0.75}{
\begin{tabular}{cccc}
      \toprule
      & BERT Sim. ($\uparrow$) & Glove Sim. ($\uparrow$) & ED ($\downarrow$) \\
      \midrule
      Syntax & 0.90 & 0.98 & 0.53 \\
      Voice & 0.87 & 0.96 & 0.48 \\
      Modifier & 0.86 & 0.90 & 0.45 \\
      Accent & 0.89 & 0.93 & 0.64 \\
      Tone & 0.81 & 0.85 & 0.41 \\
      \bottomrule
    \end{tabular}%
    \label{tab:more_semantic_sim_1}
}
\end{minipage}

\vspace{-0.6cm}
\end{table*}

\subsection{More Discussions on Dataset Diversity}
\begin{figure*}[htbp]
  \centering
  \begin{subfigure}{0.32\linewidth}
    \includegraphics[width=1\linewidth]{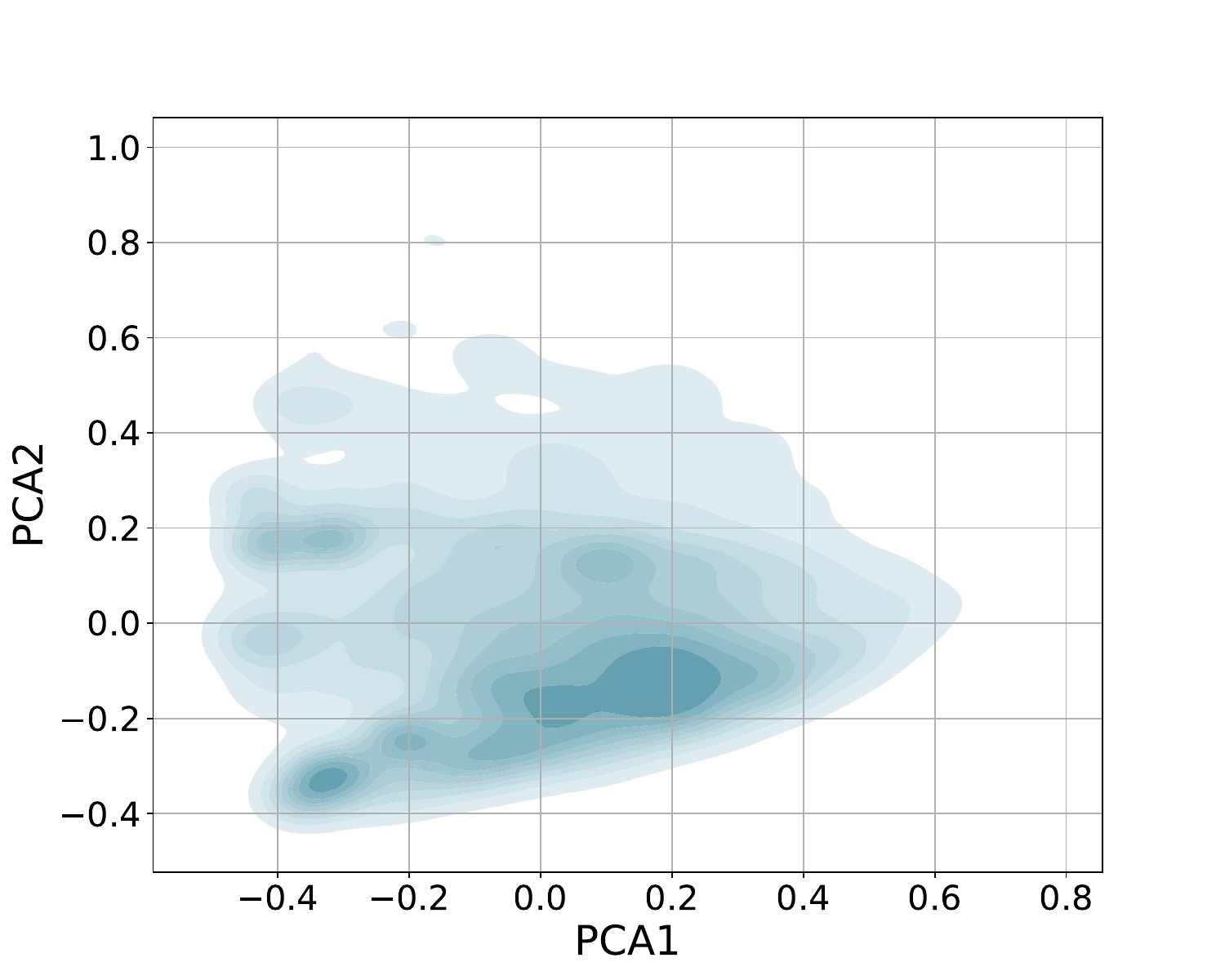}
    \caption{Original NR3D~\cite{chen2020scanrefer}}
    \label{fig:NR_density_og}
  \end{subfigure}
  \hfill
  \begin{subfigure}{0.32\linewidth}
    \includegraphics[width=1\linewidth]{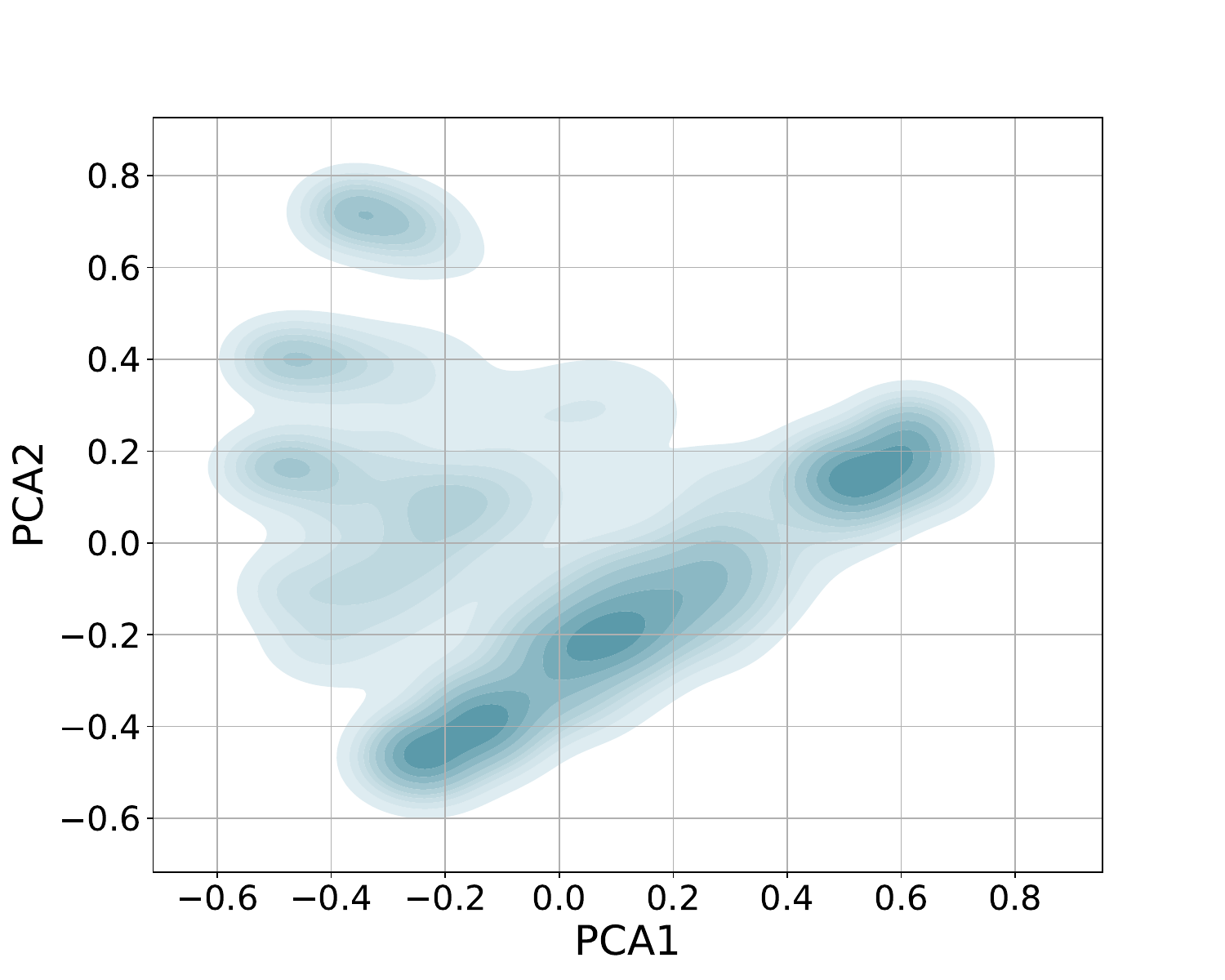}
    \caption{Original ScanQA~\cite{azuma2022scanqa}}
    \label{fig:qa_density_og}
  \end{subfigure}
\hfill
  \begin{subfigure}{0.32\linewidth}
    \includegraphics[width=1\linewidth]{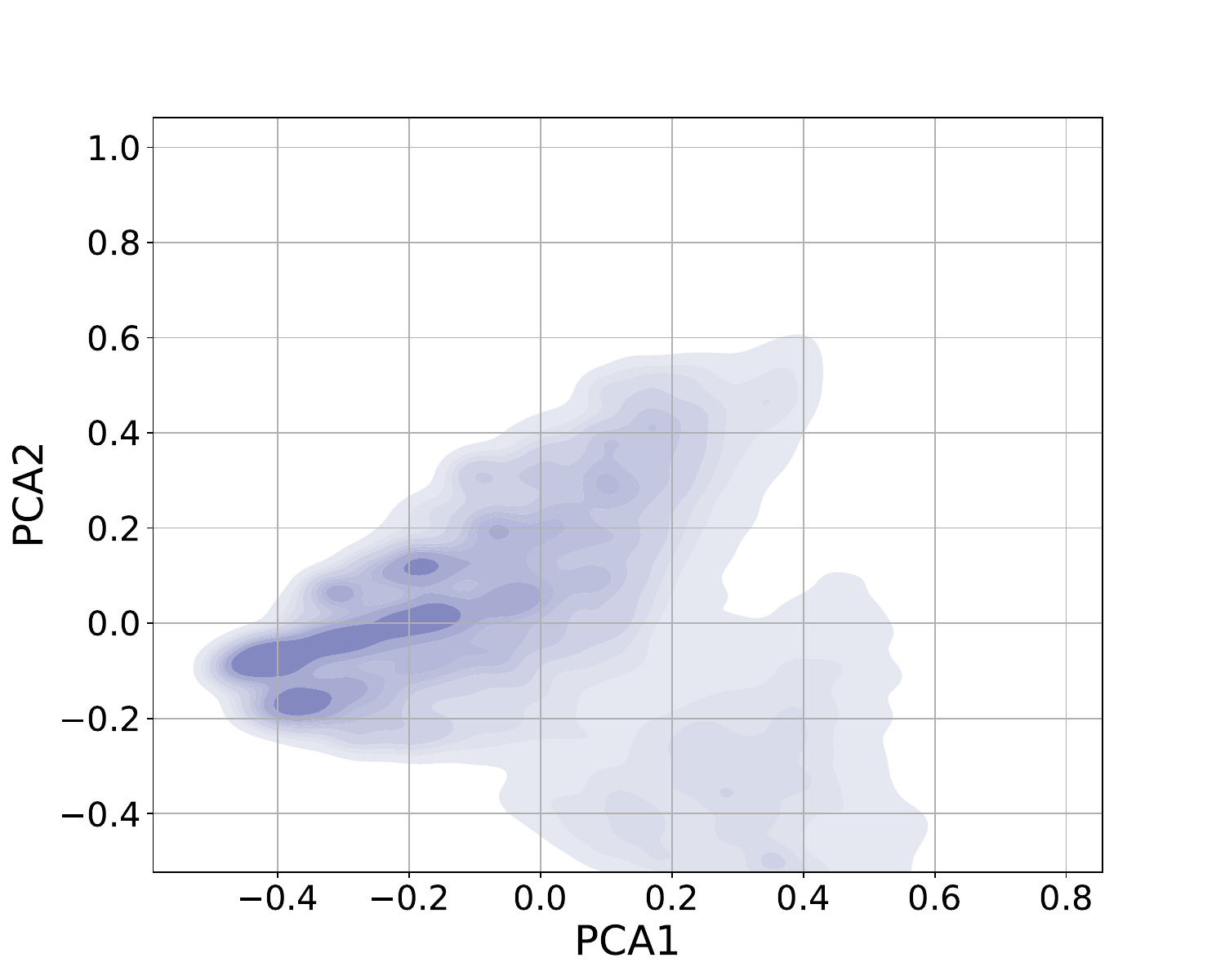}
    \caption{COCO Caption~\cite{chen2015microsoft}}
    \label{fig:NR_density_coco}
  \end{subfigure}
  \begin{subfigure}{0.32\linewidth}
    \includegraphics[width=1\linewidth]{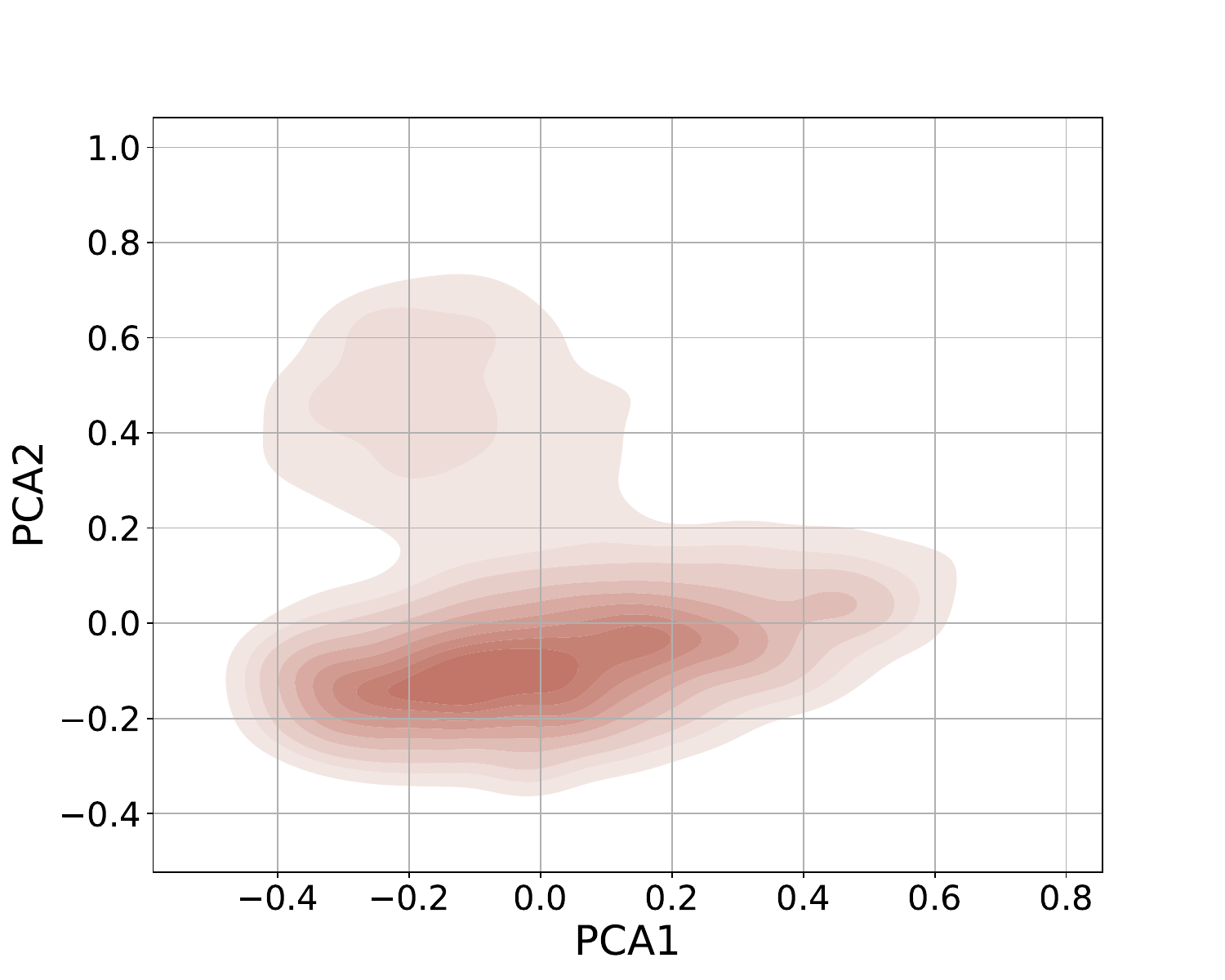}
    \caption{NR3D-R}
    \label{fig:NR_density_tone}
  \end{subfigure}
  \hfill
  \begin{subfigure}{0.32\linewidth}
    \includegraphics[width=1\linewidth]{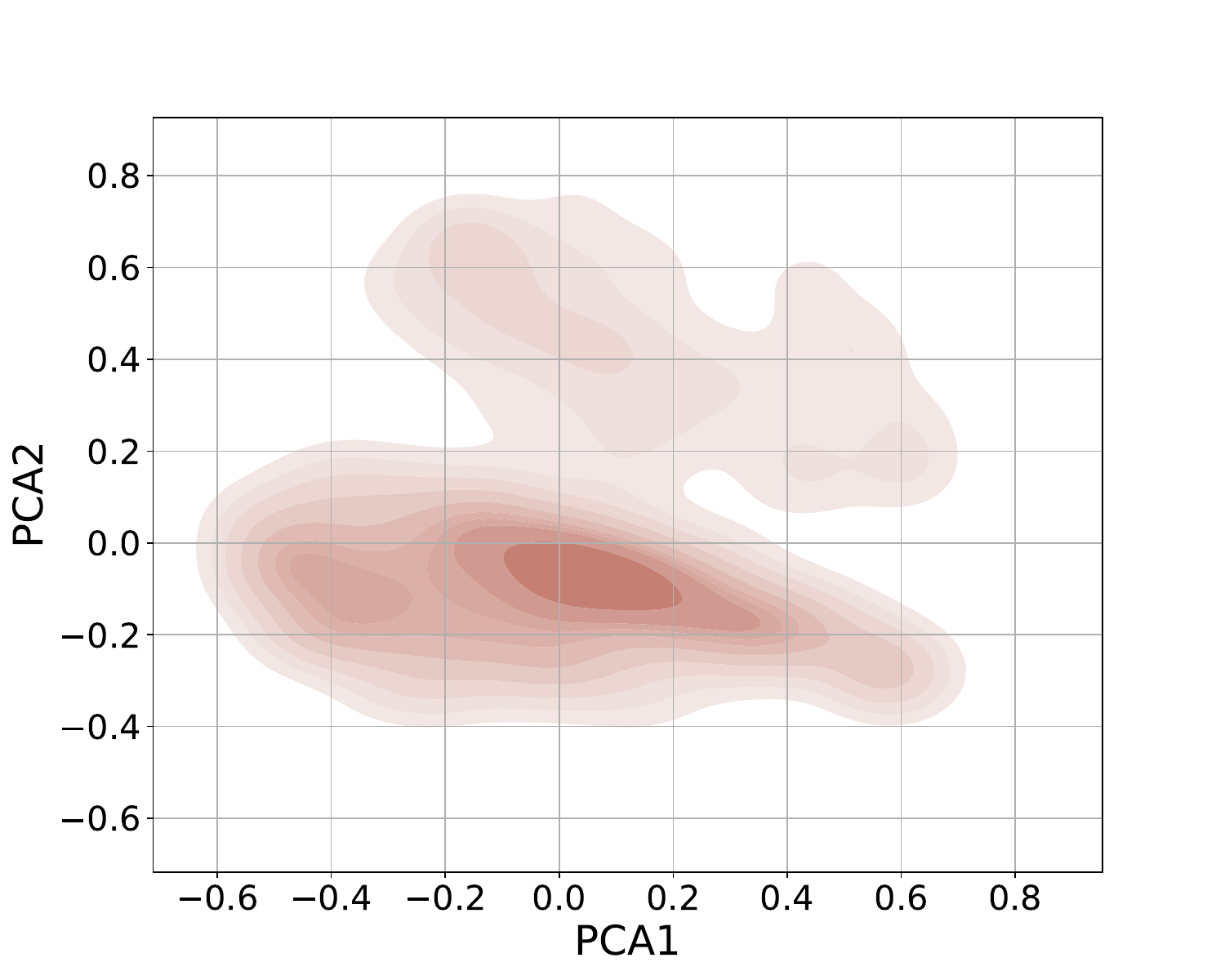}
    \caption{ScanQA-R}
    \label{fig:qa_density_tonr}
  \end{subfigure}
  \begin{subfigure}{0.32\linewidth}
    \includegraphics[width=1\linewidth]{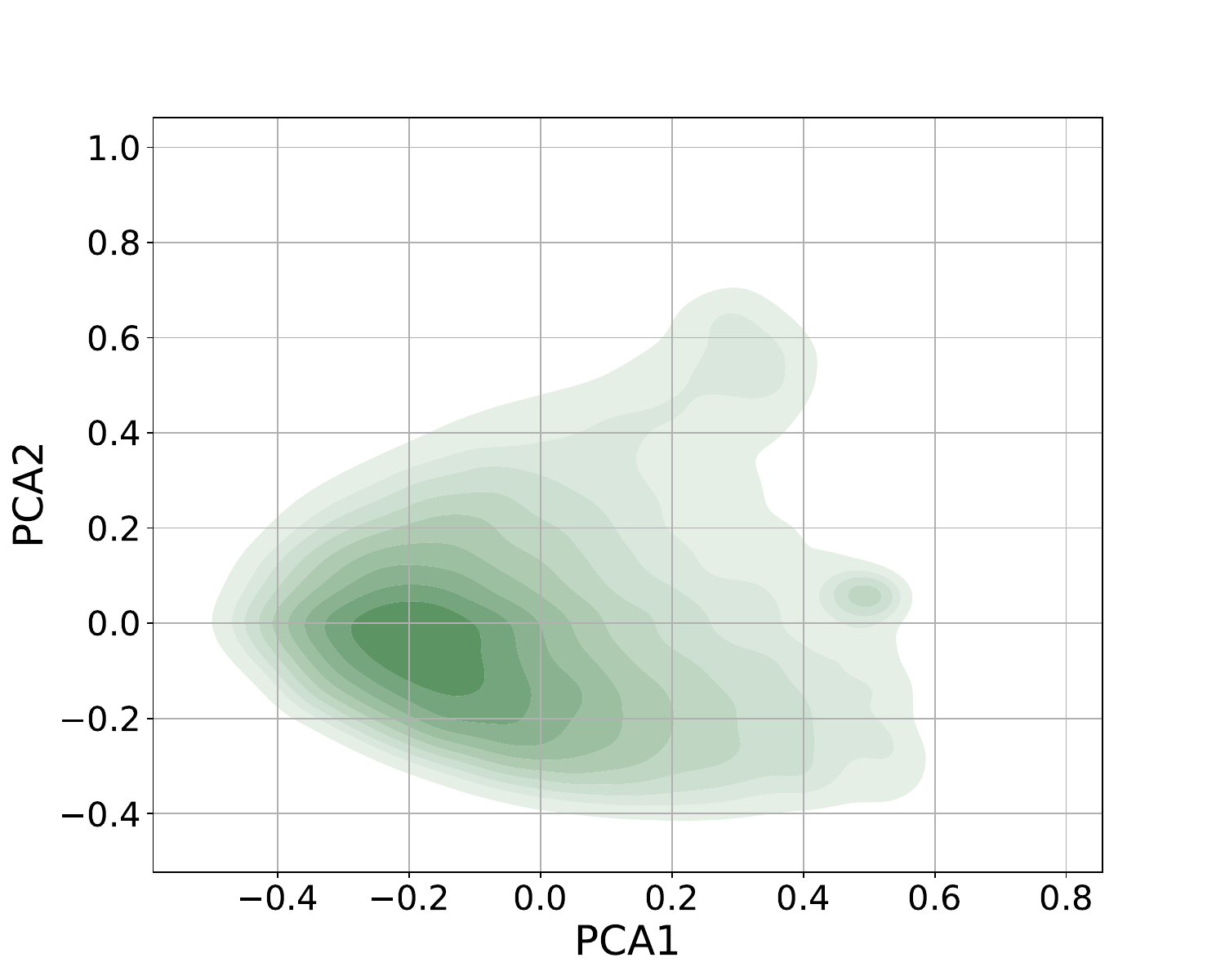}
    \caption{Open Assistant~\cite{kopf2023openassistant}}
    \label{fig:NR_density_assiatnt}
  \end{subfigure}
  \caption{Density map of datasets' vectorized syntax structure principal features. Comparing \textit{NR3D-R} and \textit{ScanQA-R} with other representative datasets.}
  \label{fig:densit_NR3D}

\end{figure*}

Fig.~\ref{fig:densit_NR3D} displays the syntax diversity of the original NR3D~\cite{achlioptas2020referit3d}, ScanQA~\cite{azuma2022scanqa}, our proposed 3D-LR dataset, and other datasets~\cite{chen2015microsoft,kopf2023openassistant} for comparisons, utilizing vectorized syntax trees. Specifically, we employ the NLTK~\cite{bird2009natural} to construct syntax trees, which represent the syntactic structure of sentences. These trees are then transformed into strings. Subsequently, we apply the Term Frequency-Inverse Document Frequency (TF-IDF)~\cite{salton1988term} technique, using TfidfVectorizer, to convert these syntax strings into quantifiable features. Following this, we implement Principal Component Analysis (PCA)~\cite{pearson1901liii} to project these features into a lower-dimensional space, enabling us to plot the density map for each dataset. In these maps, smaller dark areas signify more uniform or similar sentence patterns, indicating a lack of diversity in sentence structures within the dataset. Comparing Fig.~\ref{fig:densit_NR3D}(d) to Fig.~\ref{fig:densit_NR3D}(f), which represents a large-scale, fully human-annotated dataset, namely Open Assistant~\cite{kopf2023openassistant}, we see a similar pattern. This similarity suggests that our proposed \textit{NR3D-R} dataset accurately reflects the characteristics of natural language. On ScanQA~\cite{azuma2022scanqa}, the original dataset shows a strong, compact pattern (Fig.~\ref{fig:densit_NR3D}(b)), which model is easy to utilize such shortcut, while our \textit{ScanQA-R} (Fig.~\ref{fig:densit_NR3D}(e)) improves the diversity and makes the pattern close to natural language in real-world communication.

\begin{figure}[htbp]
\centering
\fbox{\begin{minipage}{10.5 cm}
{\small  
\texttt{[Role and Task]} \\
\texttt{You are a linguist. Your job is to assist me in paraphrasing the input query according to my needs while preserving its meaning and critical elements. [\textbf{Style Requirement:} e.g., mimic the accent of an English Speaker from a different region.]]}\\

\texttt{[Rules]}\\
\texttt{\begin{enumerate}
    \item Convert the given sentence into a more relaxed, conversational tone.
    \item Maintain the original meaning without altering it.
    \item Retain essential elements such as objects, attributes, relationships, and keywords.
    \item Present the revised sentence in JSON format, using the key "new\_sentence" for the output.
\end{enumerate}}

\texttt{[Example]}\\
\texttt{\#example 1}

\texttt{sentence: the dark blue pillow on the papasan chair}

\texttt{return answer: \{\{}

    \texttt{"new\_sentence": "The dark blue pillow resting upon the Papasan chair."}
    
\texttt{\}\}}

\texttt{\#example 2 ...}

\texttt{\#example 3 ...}\\

\texttt{[Detail Format Instruction]}\\
\texttt{You should ONLY return the JSON dictionary.}

\texttt{Python must be able to parse the response into JSON.}\\

\texttt{\#Begin Task}\\\\
\texttt{The sentence: <\{\textbf{sentence}\}>}\\}
\end{minipage}}
\caption{Overview of our prompt. To avoid complexity, we provide an example showcasing the overall structure of our prompt designed for generating sentence variants. All prompts we used, the dataset generation code and pre-generated dataset will be released later.}
\label{fig:prompt}
\end{figure}

\begin{figure}[h]
  \centering
  
   \includegraphics[width=0.9\linewidth]{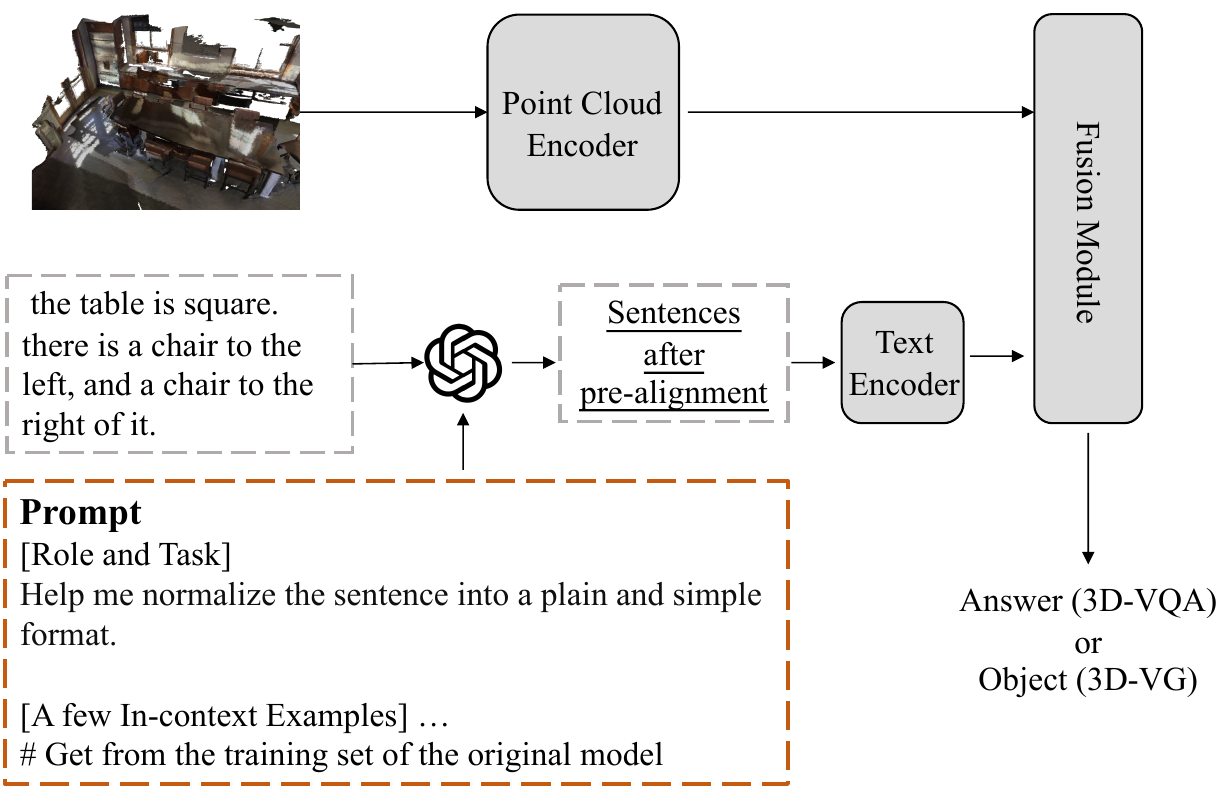}

   \caption{Overview of inference pipeline with our pre-alignment module. The entire pipeline is frozen without any training.}
   \label{fig:method}
\end{figure}

\section{Detailed Method Explanations}\label{sec:method_details}

\noindent\textbf{Pipeline of Our Pre-Alignment Module}
We introduce a method that leverages a large language model to enhance the performance of existing trained models on various language variants without necessitating re-training. This model aligns the sentence style with the training data of the original model. Specifically, it normalizes the input sentence to match the style of the data on which the models were initially trained. Fig.~\ref{fig:method} outlines the overall pipeline.

\section{More Metrics of Systematic Evaluation}\label{sec:eva_details}
\noindent\textbf{Bilingual Evaluation Understudy (BLEU Score)} 
The BLEU score (Bilingual Evaluation Understudy) was initially proposed to assess the quality of machine translation systems~\cite{papineni2002bleu}. It quantifies the similarity between a language model's translated output and the reference ground truth sentence. In the context of 3D Visual Question Answering, where models generate words, phrases, or short sentences as responses, the BLEU score evaluates the closeness of these model predictions to the correct answers. This metric effectively measures the linguistic accuracy of answers in VQA tasks, especially for 3D-LLM~\cite{hong20233d}, which generates answer token by token rather than selecting from a candidate set.

BLEU-1 is a specific variant of the BLEU score that evaluates the precision of unigrams (individual words) in machine-generated text. BLEU-1 thus quantifies how many unigrams in the generated text accurately match those in the reference text, providing a metric for lexical accuracy in tasks like machine translation or automated content generation. A higher BLEU-1 score means that the answer is more accurate. For more details, please refer to ~\cite{papineni2002bleu}.

\noindent\textbf{CIDEr Score} The CIDEr~\cite{vedantam2015cider} score originally evaluates the quality of image captions generated by computers. It calculates this score by comparing a generated sentence to a set of reference ground truth (GT) sentences. The key aspect of this comparison involves assessing the overlap of words and phrases between the generated caption and the GTs. This assessment is refined by weighting the n-grams (a contiguous sequence of $n$ items from a given sample of text) using TF-IDF, a technique that evaluates how frequently a word appears in the document relative to its commonness across all documents. The scoring favors captions that match the GTs closely and use terms specific to the given image context rather than generic words applicable to various images. Thus, in 3D-VQA, a higher CIDEr score indicates that the answer is accurate and contextually relevant.

\section{More Experimental Results}
We present additional results, including the BLEU-1 score and CIDEr score for the 3D-VQA task, in \cref{tab:bleu-table}. These metrics further elucidate the performance of 3D-LLM~\cite{hong20233d}, a generative model that produces answers token by token.
\label{sec:result}

\begin{table*}[htbp]
\centering
\resizebox{0.9\textwidth}{!}{%
\begin{tabular}{ccccccccccc}
\hline
\multicolumn{1}{c|}{\multirow{2}{*}{Method}} & \multicolumn{2}{c}{Syntax} & \multicolumn{2}{c}{Voice} & \multicolumn{2}{c}{Modifier} & \multicolumn{2}{c}{Accent} & \multicolumn{2}{c}{Tone} \\
\multicolumn{1}{c|}{}                        & BLEU-1       & CIDEr       & BLEU-1       & CIDEr      & BLEU-1          & CIDEr         & BLEU-1       & CIDEr       & BLEU-1      & CIDEr      \\ \hline
\rowcolor{mygray}
\multicolumn{11}{c}{ORACLE: (BLEU-1 = 29.62        CIDEr = 60.42)}                                                                                                                              \\ \hline
\multicolumn{1}{c|}{ScanQA}                  & 21.18        & 44.16       & 26.98        & 53.06      & 26.55           & 52.89         & 22.51        & 44.16       & 22.83       & 48.21      \\
\multicolumn{1}{c|}{w. ours}                  & \textbf{29.52}        & \textbf{60.13}       & \textbf{28.79}        & \textbf{56.93}      & \textbf{28.32}           & \textbf{56.62}         & \textbf{26.89}        & \textbf{55.53}       & \textbf{27.50}       & \textbf{57.07}      \\ \hline
\rowcolor{mygray}
\multicolumn{11}{c}{ORACLE: (BLEU-1 = 23.13        CIDEr = 37.37)}                                                                                                                              \\ \hline
\multicolumn{1}{c|}{3D-LLM*}                 & 20.19        & 33.55       & 21.97        & 33.85      & 22.41           & 38.03         & 15.45        & 32.03       & 18.52       & 28.05      \\
\multicolumn{1}{c|}{w. ours*}                 & \textbf{23.42}        & \textbf{37.48}       & 20.90        & 33.63      & \textbf{22.57}           & 36.18         & \textbf{21.37}        & \textbf{33.44}       & \textbf{21.82}       & \textbf{34.55}      \\ \hline
\rowcolor{mygray}
\multicolumn{11}{c}{ORACLE: (BLEU-1 = 37.22        CIDEr = 74.00)}                                                                                                                              \\ \hline
\multicolumn{1}{c|}{3D-LLM}                  & 38.20        & 74.21       & 36.52        & 70.79      & 36.06           & 70.00         & 35.05        & 68.28       & 22.73       & 50.85      \\
\multicolumn{1}{c|}{w. ours}                  & 37.30        & 72.94       & 36.07        & 69.94      & \textbf{36.85}           & \textbf{72.00}         & \textbf{35.71}        & 68.15       & \textbf{36.74}       & \textbf{71.65}      \\ \hline
\end{tabular}%
}
\caption{Evaluating ScanQA and 3D-LLM (a pre-trained 3D vision language model), with/without our pre-alignment module on ScanQA, a 3D-VQA task. 3D-LLM* indicates the model before fine-tuning (FT) on ScanQA, while 3D-LLM shows post-FT results.}
\label{tab:bleu-table}
\end{table*}
\subsection{More Discussions on 3D-LLM}
3D-LLM~\cite{hong20233d} is a recently proposed large scale 3D pre-training Vision-Language Model. The key idea is to use various 3D encoders to build unified 3D features. These features are first mapped into the same feature spaces as 2D images and then further used to train 2D VLMs. These 2D-VL models contain a powerful language model as a text encoder, e.g., Flan-T5~\cite{chung2022scaling}. Therefore, they have good language understanding abilities. Unlike ScanQA~\cite{azuma2022scanqa} baseline for 3D-VQA, 3D-LLM generates answer token by token. Therefore, we show the generative metrics in \cref{tab:bleu-table} to better assess the performance of 3D-LLM. 

In experimental results, ScanQA exhibited significant performance degradation across all language variants, as evidenced by BLEU-1 and CIDEr metrics. Evaluating 3D-LLM in two configurations, solely pre-trained and fine-tuned on the ScanQA task, revealed distinct outcomes. ``3D-LLM*'' indicates the model before fine-tuning (FT) on ScanQA, while ``3D-LLM'' shows post-FT results. ``3D-LLM*'', without task-specific fine-tuning, shows a performance decline in almost all variant splits except for the modifier variant. In contrast, the fine-tuned ``3D-LLM'' demonstrates greater robustness compared to the non-VLM-based ScanQA model. Notably, variations in sentence modifiers had little impact on 3D-LLM across both training setting, suggesting that its extensive pre-training may equip it to handle such language variations. However, the fine-tuned ``3D-LLM'' experienced similar performance declines across most splits. Both pre-trained and fine-tuned models showed degradation in the tone split which simulates stylistic diversity in daily communication. The fine-tuned ``3D-LLM'' exhibited more pronounced losses in the tone split than its solely pre-trained counterpart, indicating that while task-specific fine-tuning enhances downstream task performance, it may lead to catastrophic forgetting, reducing overall model robustness.

\subsection{More Discussions on Our Module}
Our pre-alignment module enhances the adaptability of trained models to different language styles without requiring re-training or data augmentations, as detailed in \cref{tab:bleu-table}. Incorporating this module into the ScanQA baseline model resulted in a performance increase of approximately 8 BLEU-1 points and 16 CIDEr points. Furthermore, with our pre-alignment, ScanQA achieved performance on par with the ORACLE across all splits. For ``3D-LLM*'', not fine-tuned on the VQA task, our method provided additional robustness, particularly noticeable in the challenging Tone split. This trend was also evident in the fine-tuned (FT) ``3D-LLM'', where our pre-alignment module effectively compensated for the loss of diversity handling capabilities after fine-tuning.

\subsection{Data Augmentation}
Section 5.3 and Tab. 4 of the main paper show that simple data augmentation, achieved by providing additional data to cover language variants for pre-training the model, still leads to unsatisfactory results. Despite the fact that a model trained on a more diverse dataset exhibits greater robustness to different language styles, its performance is worse than our straightforward, non-training-based method. Remarkably, our approach is comparable to, or even surpasses, the model trained on a dataset doubled in size, from 40,000 to 80,000 training samples. The underlying reason for this phenomenon is that while adding various styles into the training set gives the model an opportunity to learn different language variants, it simultaneously increases the learning difficulty. Consequently, the model fails to identify and overfit simple patterns within the dataset, as illustrated in Fig.~\ref{fig:densit_NR3D}. This also supports our hypothesis that existing models tend to learn shortcuts rather than achieving an actual understanding of natural language.

\section{Limitation and Broader Impact}\label{sec:limitation}
This study's limitations arise from the potential incompleteness of the 3D-LR dataset in capturing the full spectrum of natural language variations. Natural language use by humans in daily communication is complex and varies widely. Despite this challenge, we have managed to summarize five major language characteristics from linguistic theories. These were used to build a language robustness dataset, which facilitates the systematic evaluation of existing 3D-VL models and identifies their vulnerabilities. While we provided in-depth analysis to understand the reasons behind these phenomena, further studies need to focus on dataset quality, model training schemes, and the underlying causes of data augmentation failures.

This research impacts various domains, including embodied agents, autonomous navigation, robotics, and interaction with environments through language. By systematically studying and enhancing language robustness in 3D vision-language models, we offer practical benefits for applications that require the understanding of human instructions in 3D environments. The introduction of the 3D Language Robustness dataset (3D-LR) and the pre-alignment module not only demonstrates the potential for improving language robustness in 3D models but also sets a foundation for further exploration in this field.

\section{Future Works}\label{sec:future}
Future directions for this research include expanding robustness studies to other vision-language tasks and domains, such as 2D-VL, in contexts with limited data availability. A key area of investigation is understanding the low data efficiency of current data augmentation methods and discovering more efficient augmentation techniques. Another critical avenue is designing architectures that do not overfit dataset statistics and to truly understand language. This research sets the stage for a more inclusive and comprehensive understanding of 3D vision-language models.

\end{document}